\documentclass{ieeeaccess}
\usepackage{cite}
\usepackage{amsmath,amssymb,amsfonts}
\usepackage{algorithmic}
\usepackage{graphicx}
\usepackage{textcomp}

\usepackage{bm}

\usepackage[colorlinks]{hyperref}
\hypersetup{
  citecolor = {blue}
}

\usepackage{enumitem}
\usepackage[normalem]{ulem}
\useunder{\uline}{\ul}{}
\usepackage{setspace} 
\usepackage[utf8]{inputenc}
\usepackage[linesnumbered, ruled, vlined]{algorithm2e}
\usepackage{nicefrac}
\usepackage{xspace}
\usepackage{float}
\usepackage{lineno}
\usepackage{caption}
\usepackage{subcaption}
\usepackage{rotating, lscape, longtable, tabu, booktabs, siunitx, multirow}
\usepackage[capitalise, noabbrev]{cleveref}
\usepackage{verbatim}
\usepackage{mathtools}
\usepackage[table,xcdraw,dvipsnames]{xcolor}

\newcommand{\refappendix}[1]{\hyperref[#1]{Appendix~\ref*{#1}}}

\makeatletter
\AtBeginDocument{\DeclareMathVersion{bold}
	\SetSymbolFont{operators}{bold}{T1}{times}{b}{n}
	\SetSymbolFont{NewLetters}{bold}{T1}{times}{b}{it}
	\SetMathAlphabet{\mathrm}{bold}{T1}{times}{b}{n}
	\SetMathAlphabet{\mathit}{bold}{T1}{times}{b}{it}
	\SetMathAlphabet{\mathbf}{bold}{T1}{times}{b}{n}
	\SetMathAlphabet{\mathtt}{bold}{OT1}{pcr}{b}{n}
	\SetSymbolFont{symbols}{bold}{OMS}{cmsy}{b}{n}
	\renewcommand\boldmath{\@nomath\boldmath\mathversion{bold}}}
\makeatother

\def\BibTeX{{\rm B\kern-.05em{\sc i\kern-.025em b}\kern-.08em
		T\kern-.1667em\lower.7ex\hbox{E}\kern-.125emX}}

\begin{document}
	
\history{Date of publication xxxx 00, 0000, date of current version xxxx 00, 0000.}
\doi{10.1109/ACCESS.2024.0429000}

\title{Improving Token-based Object Detection with Video}
\author{
	\uppercase{Abhineet Singh}\authorrefmark{1},
	and
	\uppercase{Nilanjan Ray}\authorrefmark{2}}

\address[1]{Department of Computing Science, University of Alberta (e-mail: asingh1@ualberta.ca)}
\address[2]{Department of Computing Science, University of Alberta (e-mail: nray1@ualberta.ca)}

\markboth
{Author \headeretal: Preparation of Papers for IEEE TRANSACTIONS and JOURNALS}
{Author \headeretal: Preparation of Papers for IEEE TRANSACTIONS and JOURNALS}

\corresp{Corresponding author: Abhineet Singh (e-mail: asingh1@ualberta.ca).}

\begin{abstract}
This paper improves upon the Pix2Seq object detector \cite{p2s} by extending it for videos.
In the process, it introduces a new way to perform end-to-end video object detection that improves upon existing video detectors in two key ways.
First, by representing objects as variable-length sequences of discrete tokens, we can succinctly represent  widely varying numbers of video objects, with diverse shapes and locations, without having to inject any localization cues in the training process.
This eliminates the need to sample the space of all possible boxes that constrains conventional detectors and thus solves the dual problems of loss sparsity during training and heuristics-based postprocessing during inference.
Second, it conceptualizes and outputs the video objects as fully integrated and indivisible 3D boxes or tracklets instead of generating image-specific 2D boxes and linking these boxes together to construct the video object, as done in most conventional detectors.
This allows it to scale effortlessly with available computational resources by simply increasing the length of the video subsequence that the network takes as input, even generalizing to multi-object tracking if the subsequence can span the entire video.
We compare our video detector with the baseline Pix2Seq static detector on several datasets and demonstrate consistent improvement, although with strong signs of being bottlenecked by our limited computational resources.
We also compare it with several video detectors on UA-DETRAC
to show that it is competitive with the current state of the art even with the computational bottleneck.
We make our code and models publicly available \cite{p2sv_git}.
\end{abstract}


\titlepgskip=-21pt

\maketitle


\section{Introduction}
\label{sec:intro}
\PARstart{D}{eep} learning based object detection models conventionally produce continuous-valued and fixed-sized outputs, i.e. their output consists of real numbers rather than discrete integers and has an architecturally-determined size that is independent of the contents of the input image.
This is true of models based on both convolutional neural networks (CNNs) \cite{krizhevsky2012_imagenet} and transformers \cite{Vaswani17_transformer,Dosovitskiy21_vision_transformer,Liu2021_Swin_Transformer}.
This is not well suited to tasks like object detection and multi-object tracking (MOT) where the outputs are inherently discrete and variable-sized in nature.
In both cases, the output size depends on the input and is highly variable across inputs
since the number of objects in an image
can vary widely between images.
In addition, the space of all possible objects is
not only \textit{much} larger than the
number of objects that might realistically be present in an image
but it is also
far too large to be sampled densely by any fixed-sized representation.
This problem is greatly compounded in case of video detection and MOT
since the space of all possible objects or trajectories in a video increases exponentially with the length of the video while the actual number of such instances remains the same as in a single image.
Any fixed-sized modeling of this output space
must therefore
resort to not only sampling this space sparsely but
employing an output size that is much larger than the actual number of output instances that the model needs to produce.
An important example of this problem is provided by anchor boxes or similar methods of sampling the space of possible boxes, that are used in both RCNN \cite{Girshick15_fast_rcnn,Ren17_frcnn,Liu16_SSD,Lin17_retinanet,He2020_mask_rcnn,Cai2021_cascade_rcnn_journal} and YOLO \cite{Redmon15_yolo,Redmon17_yolov2,Redmon18_yolov3,Bochkovskiy2020_yolov4,Jocher2021ultralytics_yolov5,Wang2022_YOLOv7,Jocher2023_ultralytics_yolov8,Wang2024_YOLOv9,Terven2023_yolo_review} families of detectors.

This in turn leads to two main problems.
Firstly, it causes the loss function to become very sparse 
since the training signal only comes from objects that are actually present in the images which constitute a tiny fraction of the space of all possible objects that the network output, and therefore the loss function, needs to represent.
As a consequence, a vast majority of this output corresponds to the background and the resultant class imbalance needs to be handled through complex loss engineering.
Examples include careful domain-specific anchor box design, one-to-many ground truth (GT) to anchor box associations \cite{Ren17_frcnn}, focal loss \cite{Lin17_retinanet} and hard example mining \cite{Shrivastava2016_ohem}.
Secondly, we need to perform heuristics-based postprocessing like confidence thresholding and non-maximum suppression on the raw network output to extract the small number of discrete outputs that we actually need for downstream processing.
This introduces a disconnect between what the network learns from the training data and how it actually performs during inference.
In some cases, especially with MOT, this postprocessing can in fact have a greater impact on the overall performance than the trained model itself \cite{Bewley16_sort,Bochinski17_iou}.
Further, when the model is used as one component of a larger system that is otherwise fully differentiable, the presence of these non-differentiable heuristics makes it difficult to train the entire system end-to-end.

These issues can be largely resolved by modeling the output space with a discrete variable-length representation.
There has in fact been a trend towards such discretization in the vision literature over the last few years, whereby the outputs are represented by sequences of discrete tokens that are produced one at a time by autoregression \cite{Vaswani17_transformer}.
Pix2Seq \cite{p2s_git} is a popular language-modeling framework for computer vision that was originally introduced for object detection \cite{p2s} and then extended with a multi-task version \cite{p2s_multi} that added support for instance segmentation, keypoint detection and image captioning.
There have also been follow-up works that replace autoregression with diffusion, first for image generation and captioning \cite{p2s_diffusion}, and then for panoptic segmentation \cite{p2s_generalist}.
Other examples of tokenization in vision include SeqTrack \cite{Chen2023_SeqTrack} for single-object tracking, MMTrack \cite{Zheng2023_mmtrack} for vision-language tracking, PolyFormer \cite{liu2023_polyformer} and SeqTR \cite{zhu2022_seqtr} for referring image segmentation, and UniTab \cite{yang2022_unitab} for grounded image captioning.
There have also been related works that discretize the output space but use set prediction \cite{rezatofighi2017_deepsetnet,pineda2019_set_predict} similar to instead of language modeling. Two examples are Point2Seq \cite{Xue2022_Point2Seq} for 3D object detection, and Obj2Seq \cite{Chen2022_Obj2Seq} for pose estimation, both of which are inspired by DETR \cite{Carion2020_detr,Zhu2021_Deformable_detr}.
Finally, there have also been attempts to learn the tokenization from data instead of designing it manually, including UViM \cite{kolesnikov2022_uvim} and AiT \cite{ning2023_AiT}, both of which are multi-task models that support depth estimation and instance segmentation, among other tasks.
This trend seems to have been inspired by the remarkable success of transformer-based natural language processing (NLP) models \cite{2024_transformer_nlp_review} in recent years, especially large language models (LLMs) \cite{Zhao2023_llm_review,Minaee2024_llm_review} like Chat-GPT \cite{Radford2018_chatgpt}.

This paper proposes another step in this direction by tokenizing video object detection and potentially also MOT when future computational resources allow the video subsequence to become long enough to incorporate entire trajectories.
This single-step approach to video detection also offers an adavantage, albeit largely theoretical for now, over many existing video detectors that conceptualize video detection as a two-step problem.
These detectors first run a static detector on each individual frame and then  aggregate these frame-specific outputs to produce their video-level output.
For example, VSTAM \cite{fujitake2022_VSTAM}, PTSEFormer \cite{wang2022_ptseformer},  TransVOD \cite{zhou2022_transvod}, ClipVID \cite{Deng2023_ClipVID}, and TGBFormer \cite{Qi2025_TGBFormer} use DETR \cite{Carion2020_detr,Zhu2021_Deformable_detr} as their base detector,
while MEGA \cite{chen2020_mega}, MST-FA \cite{xu2022_mstfa}, GMLCN \cite{han2022_GMLCN}, DAFA \cite{Roh2022_DAFA}, and CFA-Net \cite{Han2022_CFANet} have Faster-RCNN \cite{Ren17_frcnn},
SparseVOD \cite{Hashmi2022_SparseVOD} and QueryProp \cite{he2022_queryprop} have Sparse RCNN \cite{Ren17_frcnn,sun2021_sparse_rcnn,Cai2021_cascade_rcnn_journal},
REPP \cite{sabater2020_repp} and YOLOV \cite{shi2023_yolov, shi2024_yolovpp} have some version of YOLO \cite{Redmon18_yolov3,ge2021_yolox},
and
SALISA \cite{ehteshami2022_salisa} has EfficientDet \cite{tan2020_efficientdet} as the base detector.
This two-step approach restricts most of these detectors to producing objects that span a predetermined and very small number of frames (usually only two) and there is no easy way to increase this number.

While our proposed detector can also be seen as an extension of the static Pix2Seq detector \cite{p2s}, it does not build upon frame-level outputs from the latter; instead, it conceptualizes and outputs the video objects as indivisible 3D boxes or tracklets that are produced by consolidating global video-level information without involving any frame-specific components.
This allows our detector to scale up to arbitrary number of frames as long as sufficient computational resources are available.
In theory, it can even generalize to MOT if the number of frames can be made large enough to span entire trajectories.

We have implemented the video detector as a part of the Pix2Seq framework \cite{p2s_git} and make our code and trained models publicly available \cite{p2sv_git} to ensure wide compatibility and ease of reproducibility.
Note that many of the tokenization concepts in this paper can be better illustrated with animations rather than still images.
Since it is not possible to include animations here, we have created a website for this project \cite{p2sv_web} which contains the videos corresponding to many of the images included here.
The links to the actual videos are included in the caption to each such image.

The remainder of this paper is organized as follows.
Sec. \ref{sec_token} first describes the tokenization strategy in Pix2Seq static detector (Sec. \ref{sec_static_token}), followed by its proposed extension for video detection (Sec. \ref{sec_vid_token}).
Sec. \ref{net_arch} provides details of the network architectures, again starting with the static image input (Sec. \ref{static_net_arch}) and followed by its adaptation for video inputs (Sec. \ref{vid_net_arch}).
Sec. \ref{sec_evaluation} details the evaluation process including datasets (Sec. \ref{sec_datasets}), metrics (Sec. \ref{sec_metrics}), training \ref{sec_training} and quantitative results (Sec. \ref{sec_results}) comparing the proposed detector with the static version as well as several state of the art video detectors from literature.
Finally, Sec. \ref{conclusions} concludes with a summary of the results and ideas for future improvements.

\section{Tokenization}
\label{sec_token}
\subsection{Static Object Detection}
\label{sec_static_token}

\begin{figure*}[!t]
	\centering
	\includegraphics[width=0.506\textwidth]{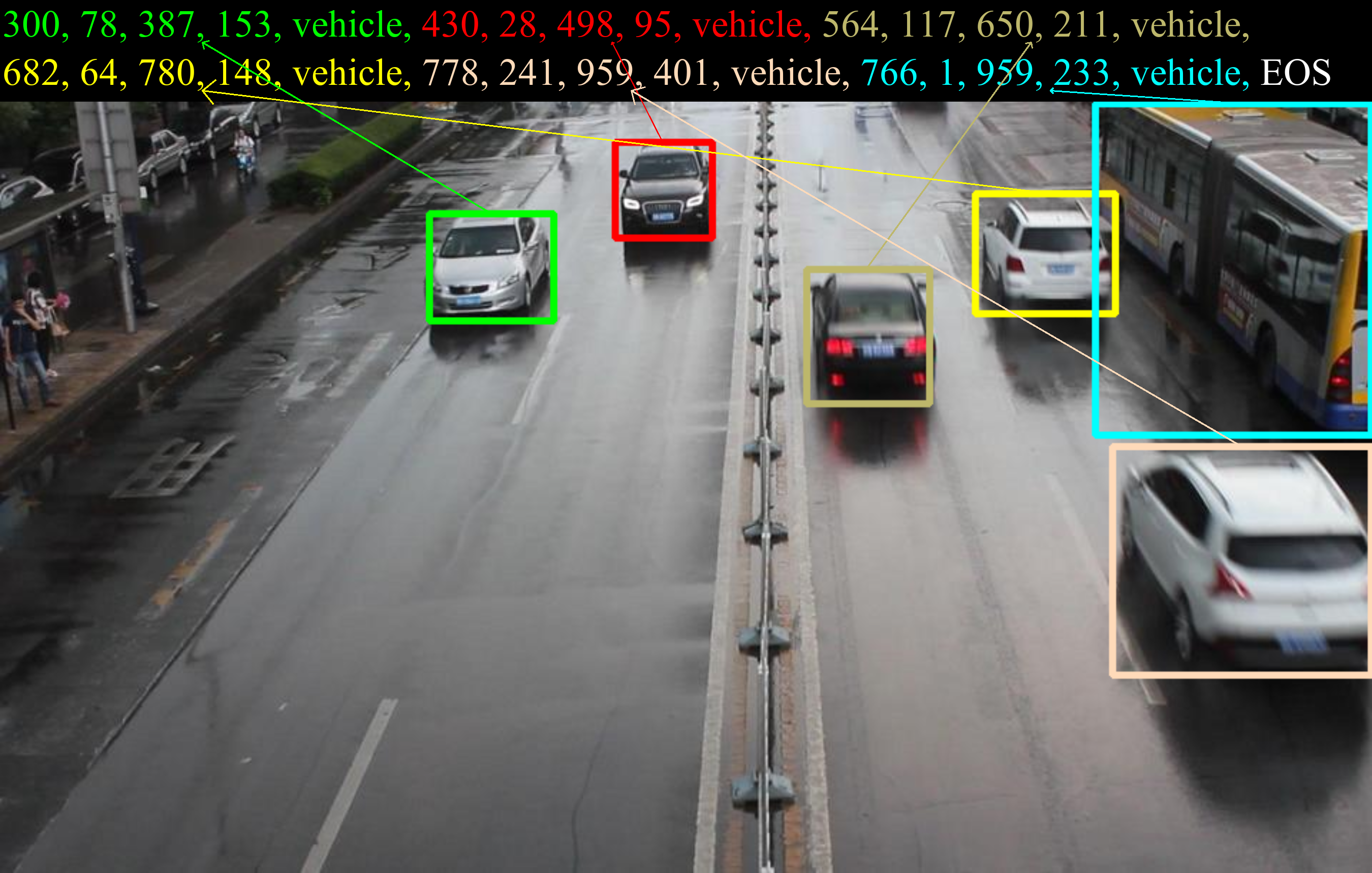}
	\includegraphics[width=0.484\textwidth]{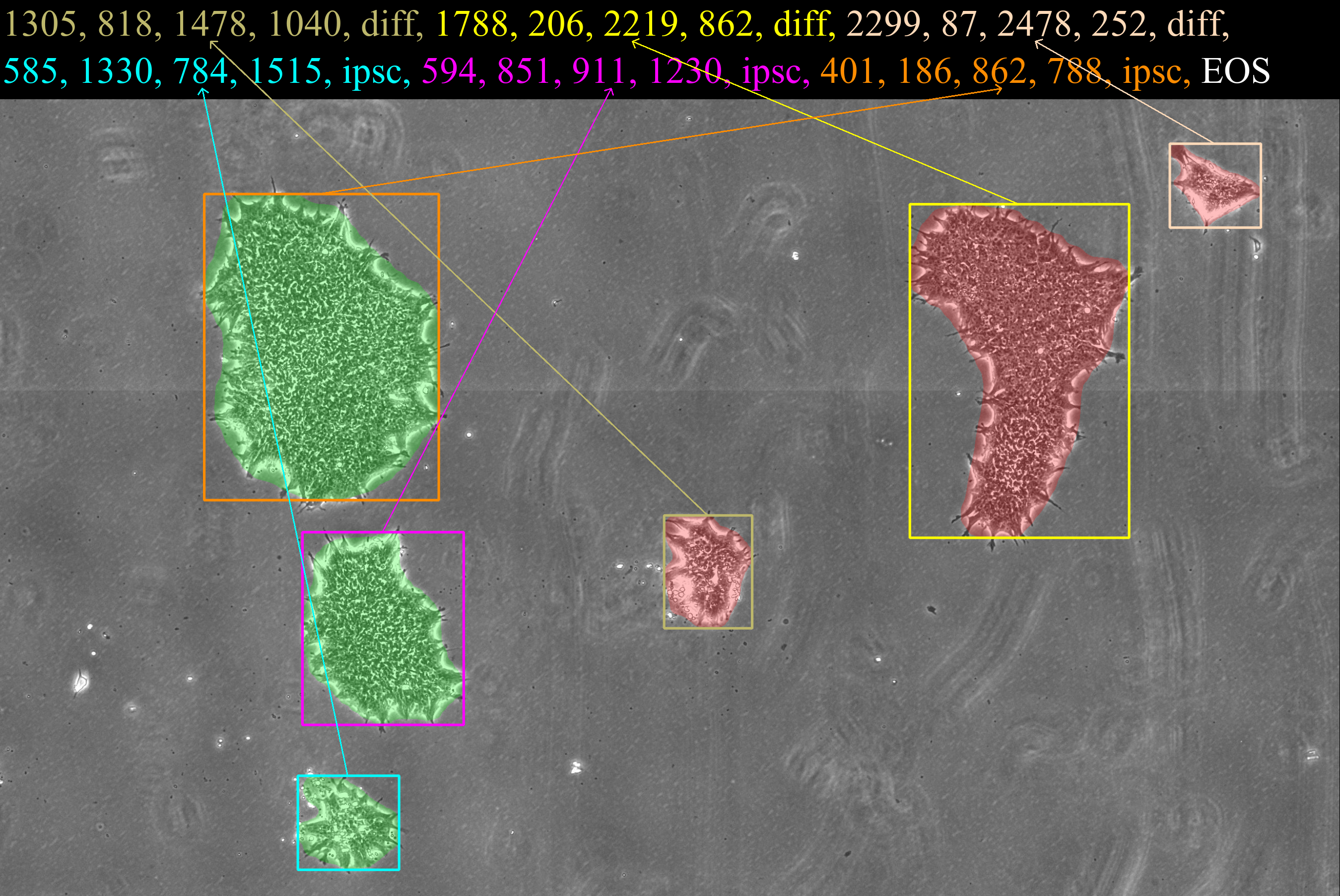}
	\caption{
		Visualization of Pix2Seq object tokenization on images from (left) UA-DETRAC \cite{Wen2020_DETRAC} and (right) IPSC \cite{Singh23_ipsc} datasets.
		UA-DETRAC has a single class, represented here by the \textit{vehicle} token, while IPSC has two classes of cells, shown here with green and red masks and represented with \textit{ipsc} and \textit{diff} tokens respectively.
		Animated versions of these images are available \href{https://webdocs.cs.ualberta.ca/~asingh1/p2s\#static_det_token_vis_detrac}{here}
		and \href{https://webdocs.cs.ualberta.ca/~asingh1/p2s\#static_det_token_vis_ipsc}{here}
		respectively.		
	}
	\label{fig:static_det_token_vis_detrac}	
\end{figure*}


Pix2Seq static detector represents each object by five tokens - four for the bounding box corner coordinates
and one for the class:
$\color{red}{\textit{ty,lx,by,rx}},\color{cyan}{\textit{cls}}$.
Here, $({\textit{lx}}, {\textit{ty}})$ and  $({\textit{rx}}, {\textit{by}})$ are respectively the top left and bottom right bounding box corners coordinates, while ${\textit{cls}}$ is the class token.
These coordinates are in image-space which is discretized into bins.
The number of such bins $H$ can be either greater than the image resolution to achieve sub-pixel accuracy or less than it to reduce the vocabulary size.
Both \textit{x} and \textit{y} coordinates are represented by the same shared set of tokens so that the total number of tokens in the vocabulary  $V$ is given by $V = H + C + r$,
where $C$ is the number of classes and  $r$ is number of reserved tokens (e.g. EOS and padding token).
Pix2Seq static detector uses $H=2K$ and $V=3K$ by default, although the framework supports $V$ upto $32K$ out-of-the-box and we have successfully trained models with $V$ upto $68K$, which seems to be limited only by the available computational resources.

The tokens for all the objects in the image are produced sequentially, so that the total number of output tokens = $5\times n + 1$, where $n$ is the number of objects in the image and the extra token at the end is the EOS token that marks the end of all objects.
The order of objects is not defined and is randomized each time the same image is shown to the network during training so the network is able to learn an order-agnostic representation of the objects.
Fig. \ref{fig:static_det_token_vis_detrac} shows examples of static detection tokenization for both single- and multi-class cases.

\subsection{Video Object Detection}
\label{sec_vid_token}

\begin{figure*}[!t]
	\centering
	\includegraphics[width=0.53\textwidth]{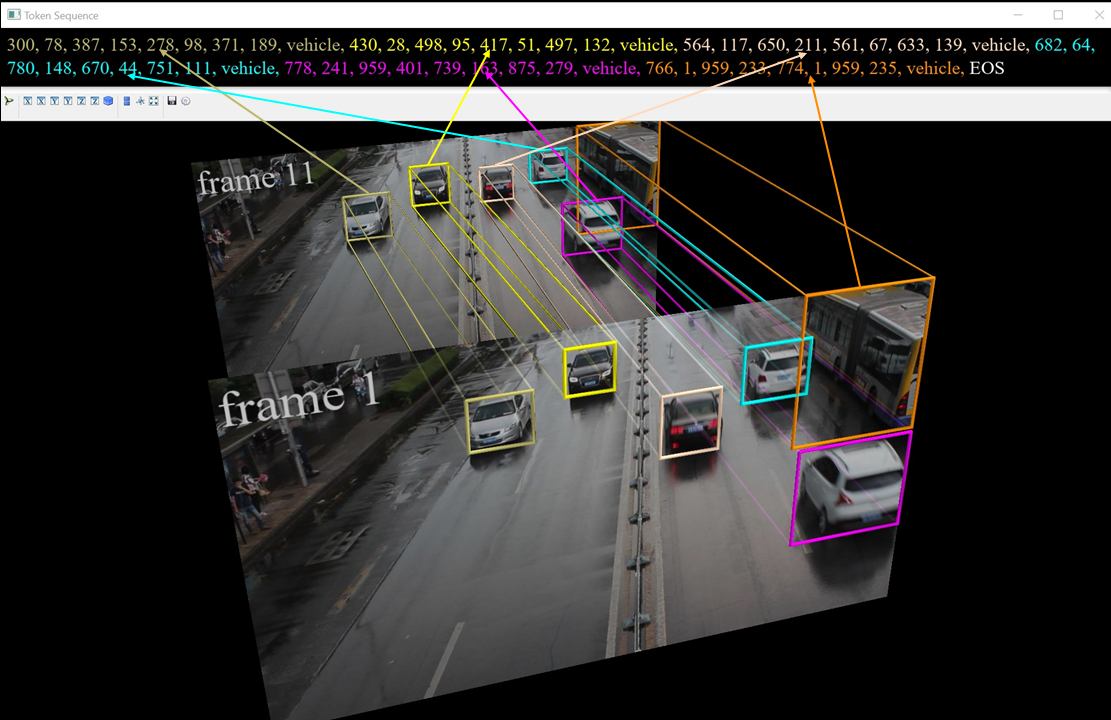}
	\includegraphics[width=0.46\textwidth]{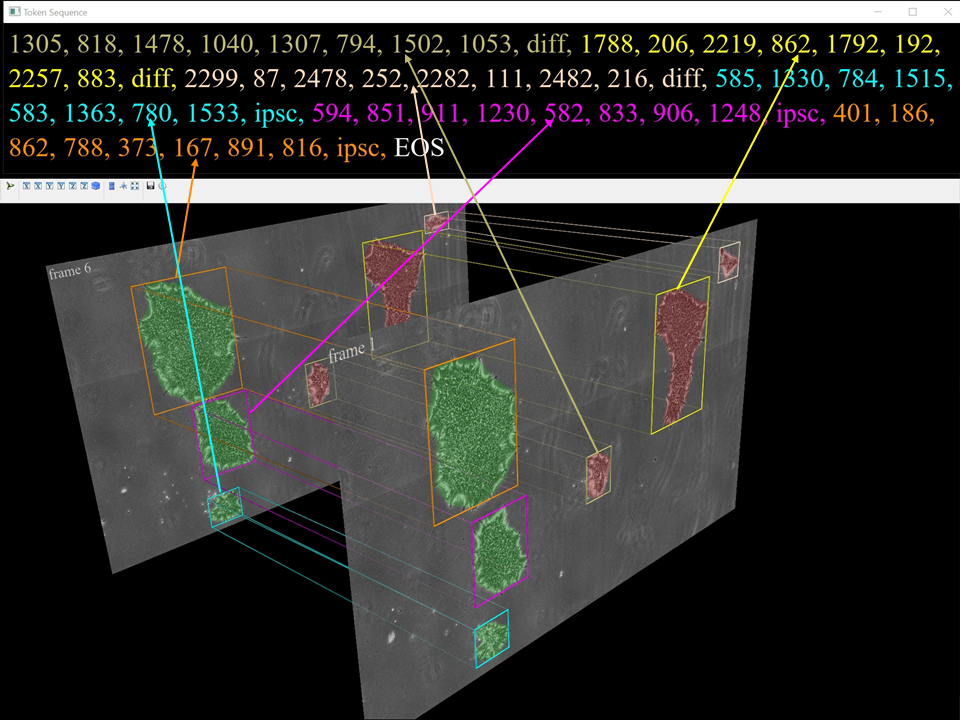}
	\caption{
		Visualization of video object tokenization on video clips from (left) UA-DETRAC and (right) IPSC datasets, both with $N=2$. UA-DETRAC shows $G=10$ while IPSC shows $G=5$.
		Animated versions of these images are available \href{https://webdocs.cs.ualberta.ca/~asingh1/p2s\#vid_det_token_vis_detrac}{here}
		and \href{https://webdocs.cs.ualberta.ca/~asingh1/p2s\#vid_det_token_vis_ipsc}{here}
		respectively.		
	}
	\label{fig:vid_det_token_vis_detrac}	
\end{figure*}


We adapted this tokenization strategy for video detection by considering all the bounding boxes of an object in an $N$-frame temporal window as a single 3D polygon or tubelet \cite{Kang16_tcnn} that can be represented by a sequence of $N$ 4-tuples, one for each bounding box, followed by the class token.
Therefore, each object can be represented by $4\times N + 1$ tokens and all $n$ objects by $n\times(4\times N + 1) + 1$ tokens.
For example, with $N=2$, an object can be represented with 9 tokens:
\\
$\color{red}{ty_1, lx_1, by_1, rx_1}, \color{ForestGreen}{ty_2, lx_2, by_2, rx_2}, \color{cyan}{cls}$
\\
Here, $(lx_i, ty_i)$ and $(rx_i, by_i)$ are respectively the top left and bottom right bounding box corner coordinates in the $i^{th}$ frame $F_i$, $1\leq i \leq N$.
Fig. \ref{fig:vid_det_token_vis_detrac} shows two examples of video detection tokenization.

\subsubsection{\textit{NA} Token}
\label{na_token}
It is possible that an object is not present in each of the $N$ frames in the temporal window.
We account for this possibility by adding a special \textit{NA} token to the vocabulary to denote non-existence.
Missing objects happen because an object either enters or leaves the scene in the middle of the temporal window, or it becomes briefly occluded in the middle of the window.
Following are examples of all three cases:
\begin{itemize}[left=0pt,topsep=0pt,noitemsep,label=\textendash]
	\item $N=3$, object leaves the scene in the third frame:
	\\
	{\small$\color{red}{ty_1, lx_1, by_1, rx_1,}
	\color{ForestGreen}{ty_2, lx_2, by_2, rx_2,}
	\color{blue}{\textit{NA,NA,NA,NA,}}
	\color{cyan}{\textit{cls}}$}
	
	\item $N=4$, object enters the scene in the third frame:
	\\
	{\small$\color{red}{\textit{NA,NA,NA,NA,} }
	\color{ForestGreen}{\textit{NA,NA,NA,NA,}}
	\color{blue}{\textit{ty}_3, \textit{lx}_3, \textit{by}_3, \textit{rx}_3, }
	\\
	\color{magenta}{\textit{ty}_4, \textit{lx}_4, \textit{by}_4, \textit{rx}_4, }
	\color{cyan}{\textit{cls}}$}
	
	\item $N=5$, object occluded in the third and fourth frames:
	\\
	{\small$\color{red}{\textit{ty}_1, \textit{lx}_1, \textit{by}_1, \textit{rx}_1,}
	\color{ForestGreen}{\textit{ty}_2, \textit{lx}_2, \textit{by}_2, \textit{rx}_2,}
	\color{BurntOrange}{\textit{NA,NA,NA,NA,}}
	\\
	\color{Purple}{\textit{NA,NA,NA,NA,}}
	\color{Brown}{\textit{ty}_5, \textit{tx}_5, \textit{by}_5, \textit{rx}_5,}
	\color{cyan}{cls}$}
\end{itemize}
\textit{NA} tokens become more common as $N$ increases, especially in scenarios where the camera field of view is limited and objects are fast-moving.
An animated example from UA-DETRAC with $N=6$ is available \href{https://webdocs.cs.ualberta.ca/~asingh1/p2s\#vid_det_token_vis_detrac_len_6}{here}.

\subsubsection{Limits on $N$}
\label{Limits_on_N}
In theory, $N$ could be large enough to cover the entire video, in which case video detection progresses into MOT.
In practice, however, it is severely limited by the amount of available RAM in the graphics processing units (GPUs) that are necessary to accelerate the training to make it time-feasible.
Using the smallest ResNet-50 based Pix2Seq architecture (Sec. \ref{static_net_arch}) and without freezing any layers, $N=16$ is the maximum that can be trained on an RTX 3090 or 4090 GPU with 24 GB RAM, which is the maximum available on a consumer-grade GPU as of this writing. 
It should be possible to get close to $N=50$ on a Tesla A100 with 80 GB RAM, which is the maximum available on a workstation GPU. 
With the backbone frozen, it is possible to go as high as $N=64$ on 24 GB RAM, though this limits the batch size to 1 per GPU or total 6 over the 3 dual-GPU servers available to us (Table \ref{tab:servers}).
This is far too small to be able to learn something as complex as the the set of all objects in a 64-frame temporal window.

We have successfully trained models upto $N=32$, although
performance drops sharply
beyond $N=8$ (Sec. \ref{exp_vid_len}), indicating that such models cannot be successfully trained on our existing hardware.
This is very likely due to the above problem of too small batch sizes,
exacerbated by the frozen backbone and the relatively small number of videos in the datasets (Table \ref{chap8_app:tab:det_dataset_stats}).
Another constraint on how large $N$ can be made in practice is the the number of training iterations required for the model to reach convergence, which increases rapidly with $N$.
Finally, the maximum sequence length $L$, that must be $\geq$ number of tokens required to represent all the objects in each temporal window, imposes another limit on $N$.
For example, with $N=64$, we need $64\times 4 + 1 = 257$ tokens to represent a single object and therefore several thousand tokens for tens of objects which are not unusual in many real-world scenarios.
Pix2Seq has a default limit of $L=512$ on the maximum number of tokens that can be produced by the network.
We have managed to increase this to as high as $L=4096$ but any increase in $L$ causes both GPU RAM consumption and training time to rise quickly, even if $N$ itself is not changed.

\subsubsection{1D Coordinate Tokens}
\label{1d_coord}
The problem of $L$ increasing rapidly with $N$ can be partially ameliorated by flattening the image-space and using the corresponding 1D coordinate tokens instead of the standard 2D $(x, y)$ tokens.
Each bounding box can then be represented by only $2$ tokens instead of $4$ and each object by $2\times N + 1$ tokens, thus reducing $L$ nearly by half.
For example, with $N=3$, an object that leaves the scene in the third frame is tokenized as:
\\
$\color{red}{tl_1, br_1,}
\color{ForestGreen}{tl_2, br_2,}
\color{blue}{\textit{NA,NA,}}
\color{cyan}{\textit{cls}}$
\\
Here, $tl_i$ and $br_i$ are respectively the flattened top left and bottom right corner coordinates of the bounding box in $F_i$.

However, using 1D coordinates greatly increases $V$ since now we need a total of $H^2$ different coordinates tokens instead of $H$ and therefore $V = H^2 + C + r$.
This severely limits the number of coordinate bins $H$ that we can use, which in turn significantly reduces the localization accuracy of the detector.
Also, while increasing $V$ does not affect the GPU memory consumption and training time to the same extent as $L$, it does cause both to increase too, thereby limiting the batch size we can use and the number of epochs we can train for.
We did perform some experiments with $H=160$ and $H=256$, using respective vocabulary sizes of $V=28K$ and $V=68K$ (Sec. \ref{1d}), but found the models to be training far too slowly to be practicable.
It appears that the 1D tokenization is too much of a change from the standard 2D version used in the pretrained weights for them to be fine-tuned successfully with the limited data and computational resources available to us.

\begin{figure*}[h]
	\centering
	\fbox{\includegraphics[width=0.477\textwidth]{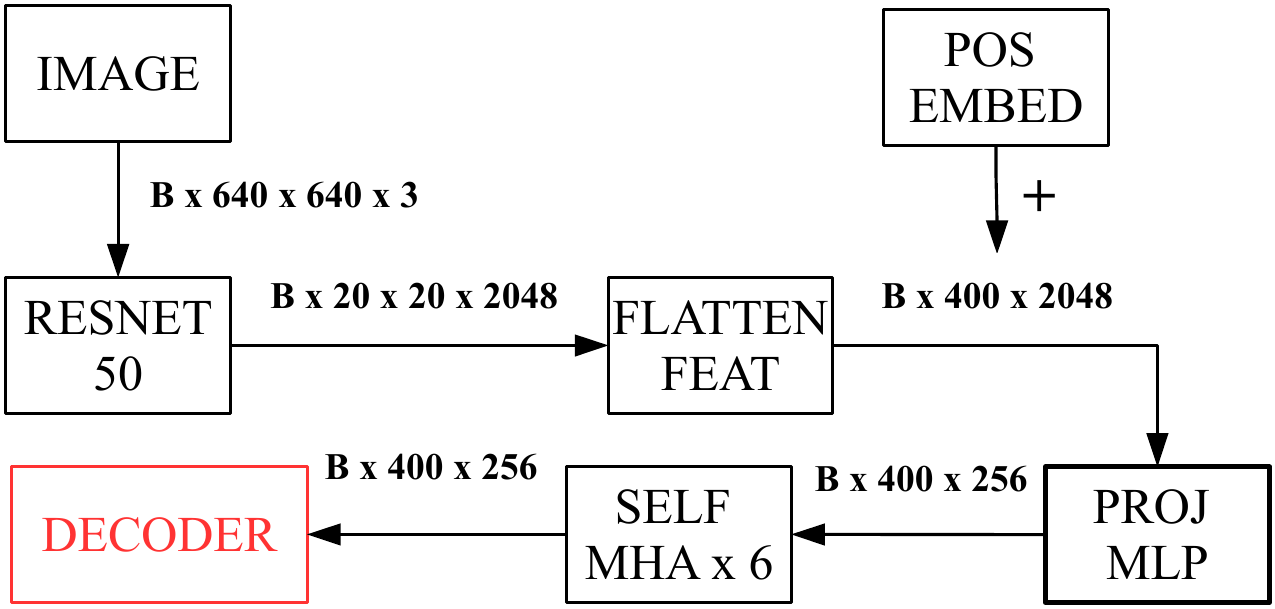}}
	\fbox{\includegraphics[width=0.463\textwidth]{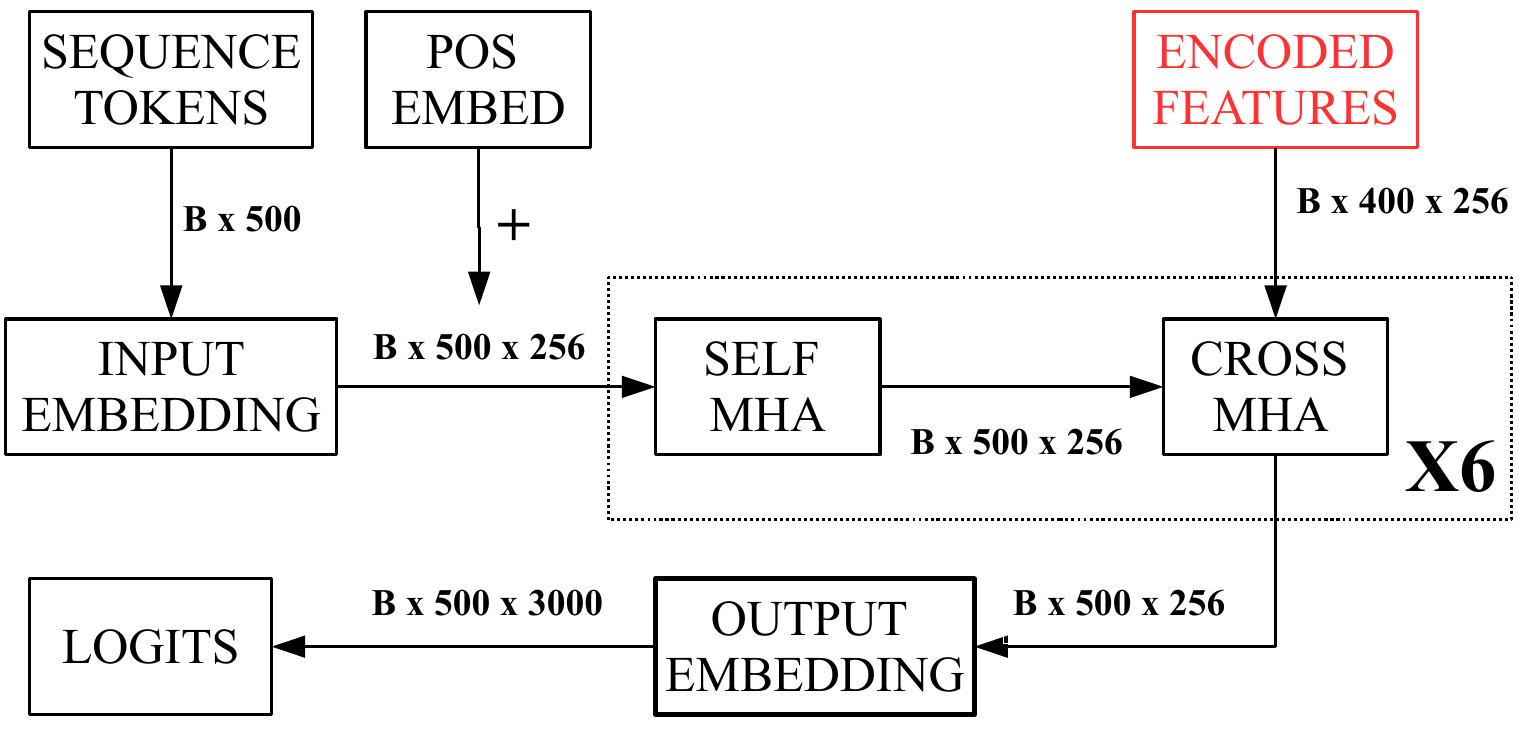}}
	\caption{
		Flow diagram representing the high-level processing in the Pix2Seq encoder on the left and decoder on the right.
		Please refer Sec. \ref{static_encoder} and \ref{static_decoder} for details of individual modules.
	}
	\label{fig:static_encoder_decoder}	
\end{figure*}

\subsubsection{Temporal Windows}
\label{Temporal_Windows}
Since we cannot process the entire video at once, we divide it into $N$-frame temporal windows which are processed one at a time in a sliding window manner.
There are two hyperparameters we can adjust to construct more varied temporal windows, both as a form of training data augmentation and to improve inference results.

The first parameter is the temporal stride $T$ which is the number of frames that separate two consecutive windows.
We can use a stride $T < N$ to introduce redundancy during inference which can help to deal with false negatives.
For example, with $N=3$ and $T=1$, we get frames 1, 2, 3 in the first temporal window; 2, 3, 4 in the second one; 3, 4, 5 in the third one and so on:
\\
$\color{red}{(F_1, F_2, F_3)}$, $\color{ForestGreen}{(F_2, F_3, F_4)}$, $\color{blue}{(F_3, F_4, F_5)}$,  $\color{magenta}{(F_4, F_5, F_6)}$, ...
\\
As a result, we have detection outputs for $F_2$ from two different temporal windows while the outputs for $F_3$ onwards come from three different windows.
For any $N$ in general, $T=1$ provides $N$-way redundancy for all but the first $N-1$ frames in the video.
These redundant outputs can be combined using non-maximum suppression to fill-in objects that might have been missed in some windows but not all of them.

The second parameter is the frame gap $G$ between successive frames in the same temporal window.
We can make $G>1$ as a form of training data augmentation to allow the network to learn to deal with faster inter-frame motions, which would be useful, for example, in handling dropped frames during live inference. 
Even for offline inference,  $G>1$ can provide us with even more overlapping temporal windows for each frame, thereby increasing the redundancy further.
For example,  with $N=6$, $T=3$ and $G=2$, we get temporal windows:
\\
$\color{red}{(F_1, F_3, F_5, F_7, F_9, F_{11})}$,
$\color{ForestGreen}{(F_4, F_6, F_8, F_{10}, F_{12}, F_{14})}$,...
\\
Another application of $G>1$ is to effectively utilize a much larger value of $N$ than can be fit in the available GPU RAM.
This can be done by training the network to predict the complete tracklet, including portions that correspond to missing frames.
For example, we can get temporal windows spanning 13 frames:
$\color{red}{(F_1, F_7, F_{13})}$,
$\color{ForestGreen}{(F_2, F_8, F_{14})}$,...
with $N=3$, $T=1$ and $G=6$.
Now, instead of only predicting portions of the tracklets corresponding to the 3 frames in the input:
\\
$\color{red}{ty_1, lx_1, by_1, rx_1}, \color{ForestGreen}{ty_7, lx_7, by_7, rx_7}, \color{blue}{ty_{13}, lx_{13}, by_{13}, rx_{13}}, \color{cyan}{cls}$
\\
we train the network to predict the entire 13-frame tracklet:
\\
$\color{red}{ty_1, lx_1, by_1, rx_1}, \color{ForestGreen}{ty_2, lx_2, by_2, rx_2}, \color{blue}{ty_{3}, lx_{3}, by_{3}, rx_{3}},...,
\\
\color{Brown}{ty_{13}, lx_{13}, by_{13}, rx_{13}}, \color{cyan}{cls}$
\\
This will require the network to be able to interpolate tracklets over the missing frames, but, as shown in Sec. \ref{exp_arch_static}, Pix2Seq seems pretty good at doing this.
All our experiments in this paper have been restricted to $G=1$ but $G>1$ provides an interesting avenue for future exploration.  
We have likewise only used $T=1$ for training, along with $T=1$ and $T=N$ for inference, and leave experiments with other values of $T$ as future work. 

\section{Network Archtectures}
\label{net_arch}
\subsection{Static Input}
\label{static_net_arch}
Pix2Seq supports two backbone architectures, namely ResNet-50 \cite{He16_resnet} and VIT \cite{Dosovitskiy21_vision_transformer}, each with three input sizes - $640\times 640$, $1024\times 1024$ and $1333\times 1333$.
We have used the smallest version - $640\times 640$ ResNet-50 - for most of our experiments so as to maximize $N$ and the training batch size $B$.
We did perform a few experiments using
$1024\times 1024$ and $1333\times 1333$ ResNet-50
as well as
$640\times 640$ VIT
but did not observe any significant performance improvements. 
Note that we have only adapted the ResNet-50 version for video modeling since the video fusion architectures proposed in Sec. \ref{vid_net_arch} are incompatible with VIT.
We leave the adaptation of VIT for video processing as future work since it requires far too much GPU memory for any video version to be practicably trainable with our computational resources.
The remainder of this section is therefore restricted to describing the ResNet-50 variant of the Pix2Seq static architecture.
The network itself has the standard transformer-based encoder-decoder architecture \cite{Vaswani17_transformer} where the encoder performs only self-attention with image features while the autoregressive decoder does self-attention with sequence features as well as cross-attention between sequence and image features. 
\subsubsection{Encoder}
\label{static_encoder}
The encoder takes the $640\times 640$ RGB image as input and applies multi-headed self-attention \cite{Vaswani17_transformer} to convert it into $400\times 256$ features which are used as input for the decoder.
This diagram in Fig. \ref{fig:static_encoder_decoder} (left) summarizes the encoder architecture.
Here,
 \begin{itemize}[left=0pt,topsep=0pt,noitemsep,label=\textendash]
	\item \textit{FLATTEN FEAT} is the spatial flatenning of the $20\times 20$ ResNet-50 feature maps into 1D feature vectors of size $400$ each
	\item \textit{POS EMBED} adds positional embedding to the flattened features
	\item \textit{PROJ MLP} is a multi-layer perceptron (MLP) that projects the $2048$ flattened feature vectors to $400\times 256$
	\item \textit{Self MHA X 6} is the multi-headed attention (MHA) operation that applies pairwise self-attention between each of the $400$ image features, which is repeated 6 times.	
\end{itemize}

\subsubsection{Decoder}
\label{static_decoder}
The decoder takes the $400\times 256$ image features from the encoder as input along with the sequence tokens.
It then applies multi-headed self-attention to the sequence embedding features, followed by cross-attention between the sequence and image features.
This self + cross MHA operation is repeated 6 times just like the self-MHA operation in the encoder.
Fig. \ref{fig:static_encoder_decoder} (right) summarizes the decoder architecture. 
Here,
\begin{itemize}[left=0pt,topsep=0pt,noitemsep,label=\textendash]
	\item \textit{SEQUENCE TOKENS} refers to the sequence of target tokens constructed from the GT objects and padded to size $L=500$
	\item \textit{INPUT EMBEDDING} is obtained by table-lookup into the $V\times 256$ weight matrix of a single linear layer, where each row corresponds to one token in the vocabulary of size $V=3000$
	\item \textit{OUTPUT EMBEDDING} is obtained by projecting the $256$-dimensional feature vectors to $V$-dimensional ones using the same linear layer whose weights are used in the input embedding module
	\item \textit{SELF MHA} applies pairwise self-attention between each of the $500$ sequence embedding features 
	\item \textit{CROSS MHA} applies pairwise cross-attention between each of the $500$ sequence features from self-MHA and $400$ image features from the encoder. 
\end{itemize}

\begin{figure}[t]
	\centering
	\includegraphics[width=0.49\textwidth]{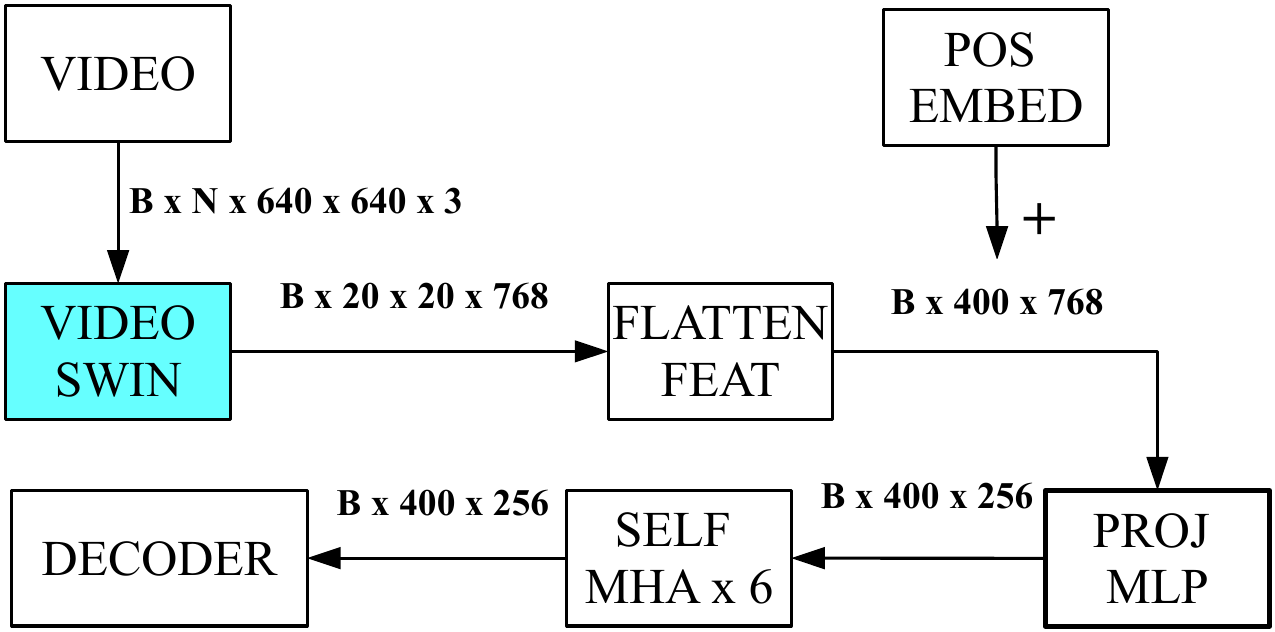}
	\caption{
		Flow diagram representing high-level processing in the early-fusion video encoder with the video Swin transformer backbone. 
		Please refer Sec. \ref{early_fusion} for details.
	}
	\label{fig:video_encoder_swin}	
\end{figure}

\subsection{Video Input}
\label{vid_net_arch}
We have determined experimentally that, like most transformer-based language models, Pix2Seq is very difficult to train from scratch on the relatively small datasets that we are working with.
Therefore, we want to be able to use as much of the pre-trained weights as possible.
This in turn requires that the baseline architecture is modified as little as possible.
With this objective, we propose three ways to adapt the architecture for processing videos.
These differ in the stage of the encoder-decoder pipeline at which the features from individual video frames are fused together. 
\subsubsection{Early Fusion}
\label{early_fusion}
This method replaces the ResNet-50 backbone with a video-specific backbone such as the video Swin transformer \cite{Liu2021_video_swin} or 3D-ResNet \cite{Feichtenhofer2018_3d_resnet} so that the feature fusion happens within the backbone itself.
We have only experimented with Video Swin Transformer so far. 
Fig. \ref{fig:video_encoder_swin} summarizes the early-fusion architecture. 
This method only changes the number of backbone feature maps from $2048$ to $768$ while the rest of the pipeline remains unchanged.
This means that we are unable to use pretrained weights only for the projection MLP that projects the flattened backbone features from $400\times 768$ to $400\times 256$.
Of course, we also lose the pretrained weights for the ResNet50 backbone itself but the video Swin transformer implementation we are using \cite{video_swin_tf_github} comes with its own weights which seem to work well enough as long as the backbone is not frozen while training.
Unlike the Pix2Seq pretrained weights which were trained for token-based object detection, this backbone was trained on the Kinetics action recognition dataset \cite{kinetics_dataset} with conventional
modeling.
As a result, the overall network fails to generalize to token-based tasks if the backbone is kept frozen (Sec. \ref{exp_fbb}).
We have experimented with both tiny and base variants of this backbone but found them to have similar performance
inspite of the latter having more than three times the number of parameters (87.64M versus 27.85M), probably because of the much smaller $B$ necessitated by this larger network size.

\subsubsection{Middle Fusion}
\label{middle_fusion}
\begin{figure}[t]
	\centering
	\includegraphics[width=0.49\textwidth]{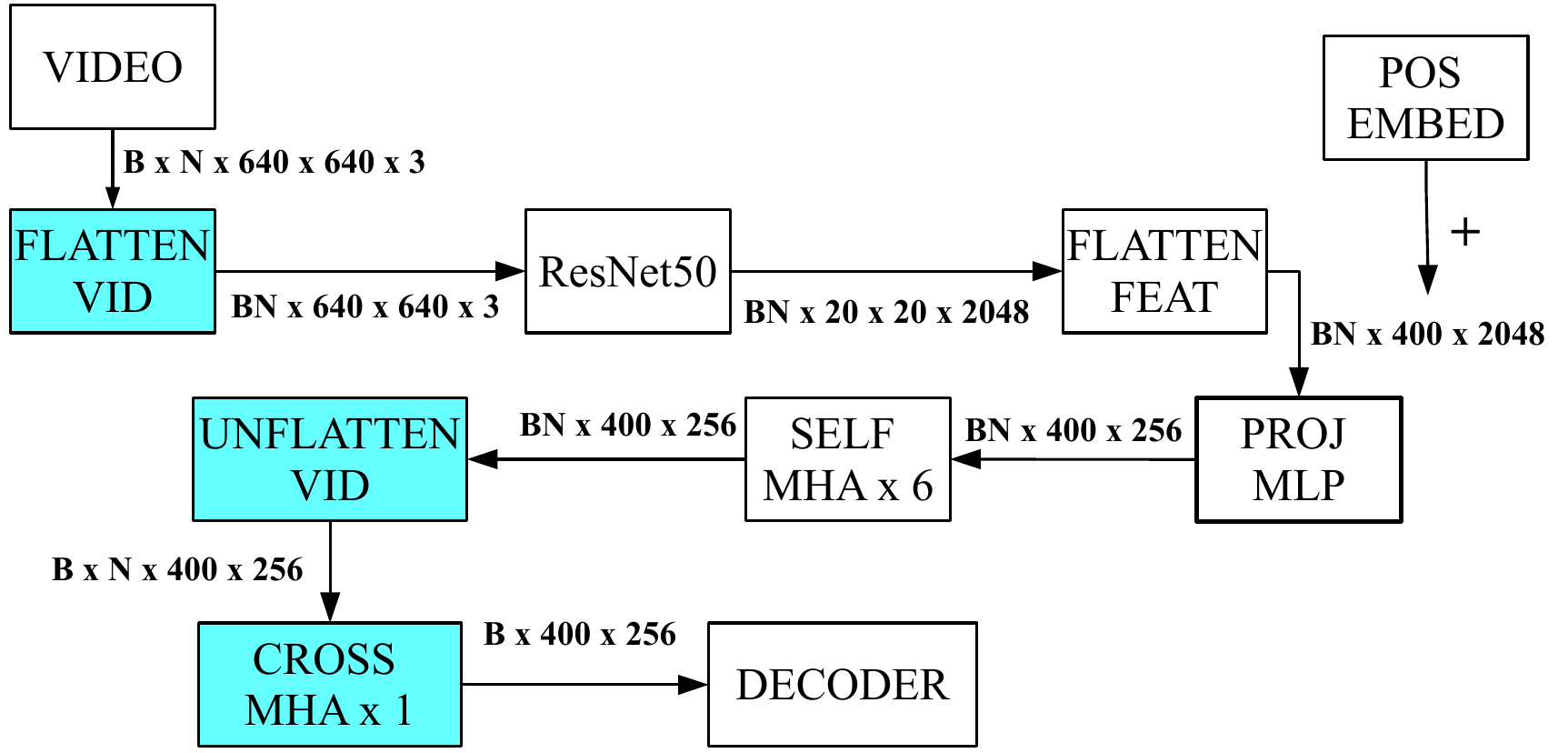}
	\caption{
		Flow diagram representing high-level processing in the middle-fusion video encoder.
		Please refer Sec. \ref{middle_fusion} for details.
	}
	\label{fig:video_encoder_middle}	
\end{figure}
As shown in Fig. \ref{fig:video_encoder_middle}, we first flatten the temporal dimension along the batch dimension to replace $B$ with $B\times N$ images (and feature maps), while leaving the rest of the encoder pipeline unchanged.
The result of this reshaping is that all the subsequent operations up to and including self-MHA are performed to each one of the video frames independently as if we had a batch size of $B\times N$ instead of $B$.
Once we have the $BN \times 400 \times 256$ output from the self-MHA module, we unflatten the temporal dimension to separate out the $400 \times 256$ features for each one of the $N$ video frames.

\begin{figure}[t]
	\centering
	\includegraphics[width=0.49\textwidth]{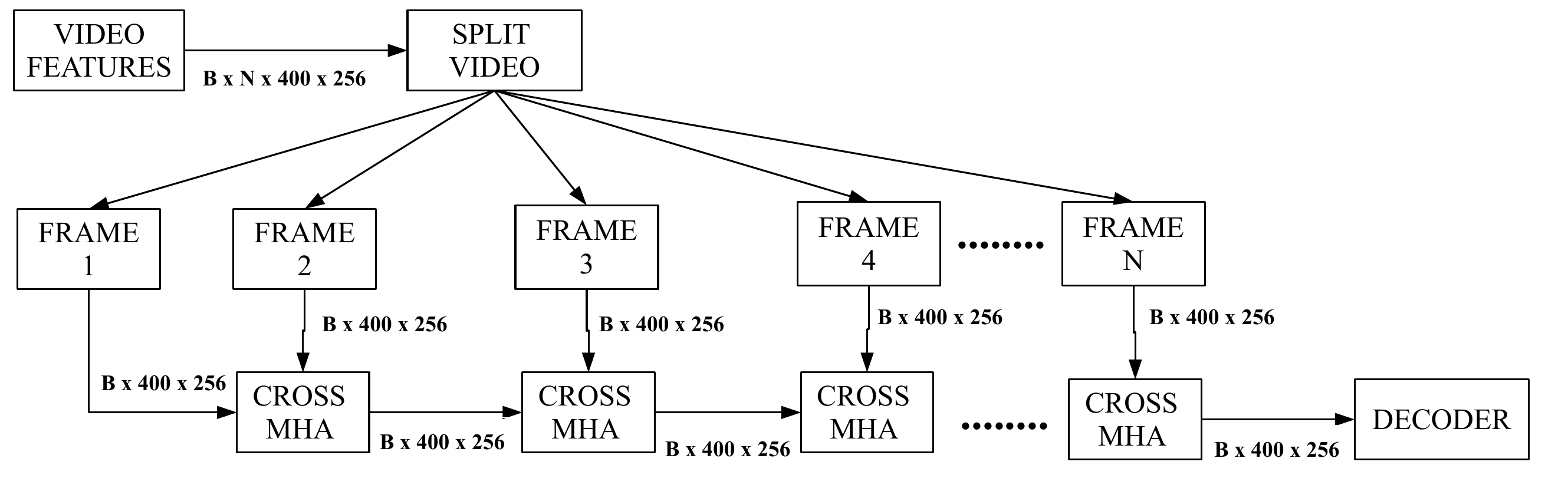}
	\caption{
		Flow diagram for the pairwise compositional variant of the cross-MHA module in the middle-fusion video encoder.
		Please refer Sec. \ref{middle_fusion} for details.
	}
	\label{fig:cross_mha}	
\end{figure}

We then apply some form of cross-attention between the features from different video frames to fuse them together.
The specific technique we have used is pairwise compositional cross-MHA as shown in Fig. \ref{fig:cross_mha}.
We first apply cross-MHA between the features of $F_1$ and $F_2$, then between the output of this operation and features of $F_3$, then between the output of this operation and features of $F_4$ and so on. 
For the sake of simplicity, we share weights between all the cross-MHA operations although it is also possible to have separate weights for each one.
Another way to perform video cross-MHA is hierarchical consecutive cross-MHA as
shown in Fig. \ref{fig:cross_mha_hierarchical}.
There are many other ways to perform this operation but these are the only two that we have tried and the pairwise compositional variant is the one we have used for all of our experiments since the hierarchical version is less stable to train.
This is likely due the excessive number of cross-MHA operations - $N\times (N-1)/2$ - that that latter requires
compared to the former which only needs $N-1$ operations.

\begin{figure*}[t]
	\centering
	\fbox{\includegraphics[width=0.52\textwidth]{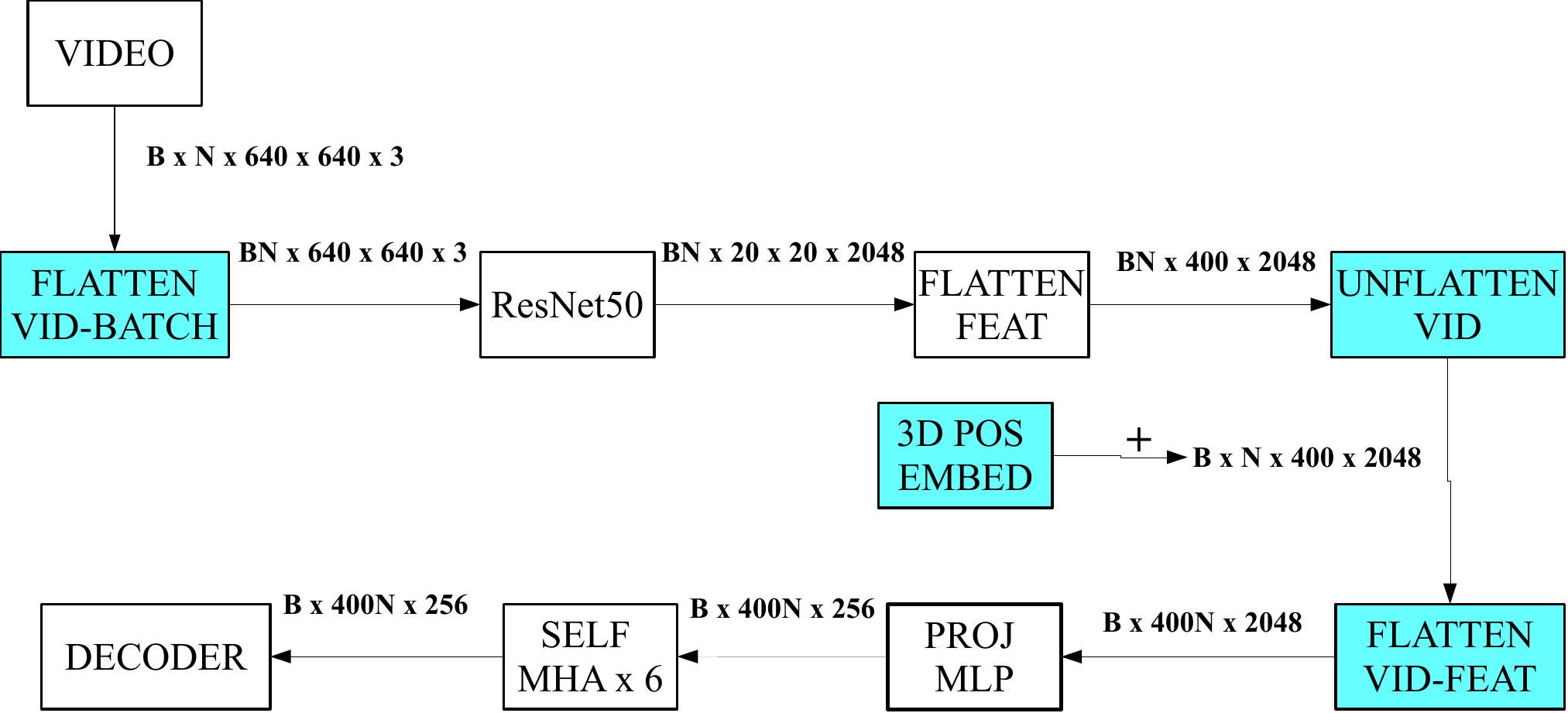}}
	\fbox{\includegraphics[width=0.44\textwidth]{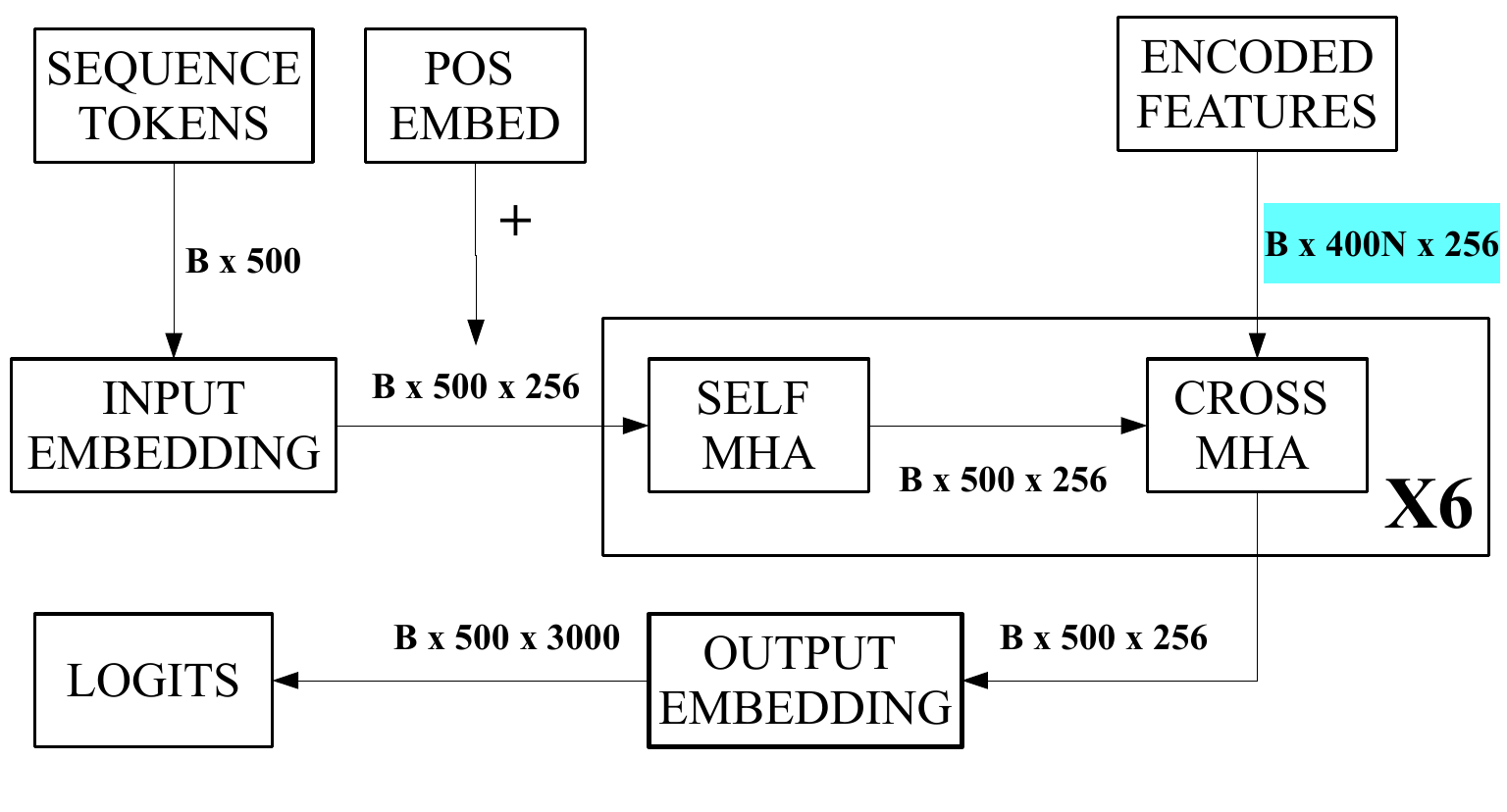}}
	\caption{
		Flow diagram representing high-level processing in the late-fusion video encoder and decoder on the left and right respectively.
		Please refer Sec. \ref{late_fusion} for details.
	}
	\label{fig:video_encoder_decoder_late}	
\end{figure*}

Note that, unline the self-MHA operation, the video cross-MHA operation cannot be repeated multiple times since the input and output shapes of each such operation are different.

Middle-fusion allows us to use all the pretrained weights but we have to learn the compositional cross-MHA weights from scratch.
This can sometimes lead to a bit of instability during training, especially for larger values of $N$.
Nevertheless, middle fusion outperforms the other two methods in most cases (Sec. \ref{exp_arch}), albeit by small margins, so this is the one we have used for most of our experiments.

\subsubsection{Late Fusion}
\label{late_fusion}
This is the only method where feature fusion happens in the decoder, unlike the other two methods where it happens in the encoder so that the decoder remains exactly the same as in the static architecture.
Fig. \ref{fig:video_encoder_decoder_late} shows the late-fusion encoder and decoder.
The encoder here is identical to middle-fusion as far as generating the $BN \times 400 \times 2048$ backbone features, after which the temporal dimension is unflattened to separate frame-specific features.
3D position embedding is then added to these to encode information about the position of each frame within the video.
The 3D position embedding parameters are the only ones for which we cannot use pre-trained weights and therefore have to learn from scratch.
The $400$ features for each of the $N$ frames in each video are then concatenated together into $400\times N$ features.
This creates an overall feature map of size $B \times 400N \times 2048$ which is processed normally for the remainder of the encoder pipeline, except that now we have $400\times N$ features instead of $400$. 
Each of these $400\times N$ image features is then cross-attended with each of the $500$ sequence features in the decoder so that every single frame is directly able to attend to every single output token. 

In theory, we would expect that the flexibility of every frame being able to directly affect every token would make it possible to train better models since visual information from any frame can be used to improve the prediction of tokens corresponding to both past and future frames.
This should be particularly useful for handling occlusions since the frame where an object is actually occluded contains little to no visual cues about this occluded state but this information can be obtained from past and future frames where the object is not occluded.
Late-fusion is also the method with the fewest parameters for which we are unable to use the Pix2Seq pretrained weights (409K in late fusion versus 800K in middle-fusion and 24M in early-fusion)
However, in practice, late-fusion has not shown any consistent performance improvement over the other two methods.

\section{Evaluation}
\label{sec_evaluation}

\subsection{Datasets}
\label{sec_datasets}
We have evaluated our detector
the folowing
three
datasets:
\begin{itemize}[left=0pt,label=\textendash]
	\item ACAMP Canadian Animal Detection (ACAD) \cite{Singh2020_Animal_det}: We have trained on only 3 of the 8 configurations \cite[Table~4]{Singh2020_Animal_det} -  $\#1$, $\#3$ and $\#4$ - since these are the only ones that contain video information.
	\item IPSC \cite{Singh23_ipsc}: We have trained on both early and late-stage training configurations  \cite[Sec.~2.3]{Singh23_ipsc} and tested both sets of models on the same test set containing the first 16 images from each sequence.
	\item UA-DETRAC \cite{Wen2020_DETRAC}: This is a class-agnostic large-scale dataset with long and diverse sequences containing relatively long-term motion information that is missing from the previous two datasets.
	This allows us to at least partially rule out dataset limitations when comparing the static Pix2Seq detector with our proposed video version, especially for larger values of $N$.
\end{itemize}
Table \ref{chap8_app:tab:det_dataset_stats} provides quantitative details for these datasets.
\begin{table}[t]
	\centering
	\caption{
		Quantitative details of the datasets used for testing video detection.
		Number of objects is the total number of frame-level objects (ignoring instance information) while the number of trajectories is a total number of video-level object instances.	
		For example, if an object enters the scene in frame 1 and leaves the scene in the frame 151 without being occluded in any frame, this will count as a single trajectory but 150 objects.
	}
	\begin{tabular}{l}
		\includegraphics[width=0.47\textwidth]{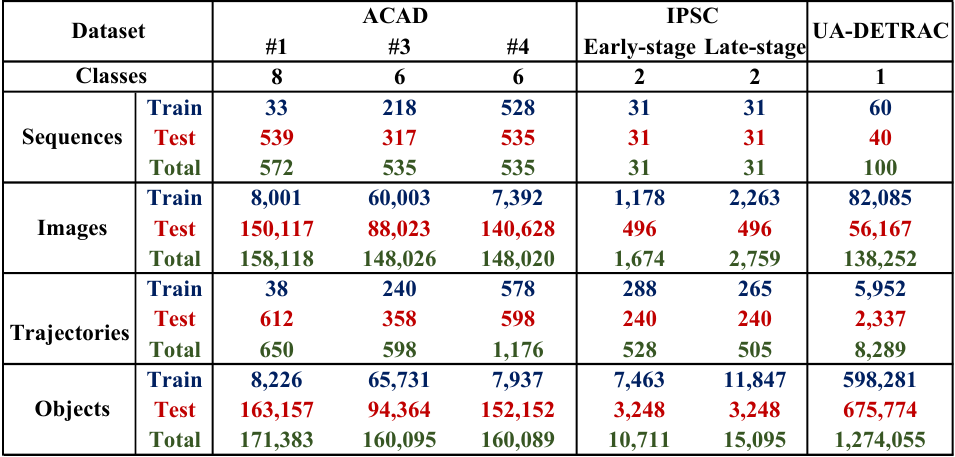}
	\end{tabular}	
	\label{chap8_app:tab:det_dataset_stats}
\end{table}

\subsection{Metrics}
\label{sec_metrics}
We have used a wide range of object detection metrics since different metrics are more suited to different application domains.
These metrics include mAP, mRP and cRP for ACAD \cite[Sec.~4.1]{Singh2020_Animal_det}, and ROC-AUC, RP-AUC, FN-DET, FP-DUP, and FP-NEX for IPSC \cite[Sec.~3.1,~3.2]{Singh23_ipsc}.
Although several of these detection metrics are strongly correlated with each other in many scenarios, we have empirically found RP-AUC to correspond best with the overall detection performance
for most practical purposes.
Our principal objective in this paper is to compare the baseline Pix2Seq static detection model with our video detection model
as well as compare between different variants of the latter, so we use RP-AUC as the single main metric for this purpose.
We do still use the domain-specific metrics proposed in \cite[Sec.~4.1]{Singh2020_Animal_det} and \cite[Sec.~3]{Singh23_ipsc} when comparing the token-based models with the conventional models from those chapters.

RP-AUC measures both localization and classification performance so we also use a class-agnostic version of RP-AUC which we have termed \textbf{cRP-AUC}.
Similar to the cRP metric in \cite[Sec.~4.1]{Singh2020_Animal_det}, cRP-AUC is computed by considering all the predicted and GT objects to belong to the same class so that misclassifications are not penalized and we are able to measure the localization performance alone.
In many practical applications,
especially those involving human-in-the-loop systems,
being able to correctly detect the presence or absence of an object is more important than classifying it correctly and this metric allows us to measure the suitability of a detector for such systems.

\subsection{Training}
\label{sec_training}
\subsubsection{Setup}
\label{sec_hw}
We have performed most of the training on four GPU servers, each with 2 $\times$ Geforce RTX 3090 24GB GPUs.
A few of the larger models were trained on a cloud server with 2 $\times$ Tesla A100 80GB GPUs.
We have also trained some of the smaller models on a fourth GPU server with 3 $\times$ Geforce GTX 1080Ti 11GB GPUs.
Finally, we used a fifth GPU server with a Geforce RTX 3090 24GB and a Geforce RTX 3060 12GB for running inference.
Note that expensive resources such as the Tesla A100 server were used for a very limited time in this study so most of the models reported here could not benefit from these.
More details of these servers are provided in Table \ref{tab:servers}.
We trained all the models with the default hyperparameter settings provided by the Pix2Seq authors, except for adjusting $B$ to the maximum that would fit on the GPUs, along with $V$ and $L$ as needed for each model configuration.

\subsubsection{Validation}
\label{sec_validation}
Pix2Seq codebase does not support performing validation as part of the training run.
While we did add support for this, we found that it not only slowed down training significantly, but also caused random crashes due to running out of GPU memory.
As a workaround, we implemented a remote validation pipeline where the inference server periodically polls the training server for new checkpoints and runs inference on the latest checkpoint thus found.
We set the time period between successive polling attempts to the maximum of two hours or the inference time.
Since inference requires much less GPU memory than training, it is possible to simultaneously perform validation for multiple training runs on the same inference server.

We performed validation directly on the test set rather than a subset of the training set, as is considered the standard practice.
We had to do this both in order to reduce the total training time as well as to utilize as many of the limited number of labeled images for training as possible.
A separate validation set is principally needed to prevent overfitting but it turned out that each of our test sets is sufficiently similar to the corresponding training set that the test performance reached a plateau in every single case and did not decline even when training was continued for several hundreds of thousands of iterations after this plateauing.
Some of the test sets
are too large to complete inference within a reasonable timeframe (e.g. $< 10$ hours)
In such cases, we validated on a small representative subset of the test set (e.g. 10 frames per sequence)
after empirically confirming
that the performance trend on this subset was consistent with the full test set.

\subsubsection{Distributed Training}
\label{sec_distributed}
Pix2Seq codebase does support multi-machine distributed training and we used it to train a few models over two or more of the four dual RTX 3090 servers in order to use the combined $96$ to $192$ GB of GPU memory.
It would have been extremely beneficial to have been able to do this for all (or at least most) of the models, especially on larger datasets like ACAD and UA-DETRAC.
However,
Pix2Seq distributed training implementation is optimized for Tensor Processing Units (TPUs) rather than GPUs and this,
combined with the relatively slow network connection between our GPU servers,
made the overhead so high that the GPU usage during these runs was $< 50\%$ nearly the entire time, and the GPUs spent a good fraction of that time idling.
In addition, two of our servers suffer from hardware incompatibility between their motherboards and RTX 3090 GPUs which causes them to restart randomly after a while when used in a distributed training setup.
These issues in turn extended the training time to such an extent (often to several weeks) that distributed training was feasible for only a few models.

\subsection{Results}
\label{sec_results}
\begin{figure}[!t]
	\centering
	\includegraphics[width=0.49\textwidth]{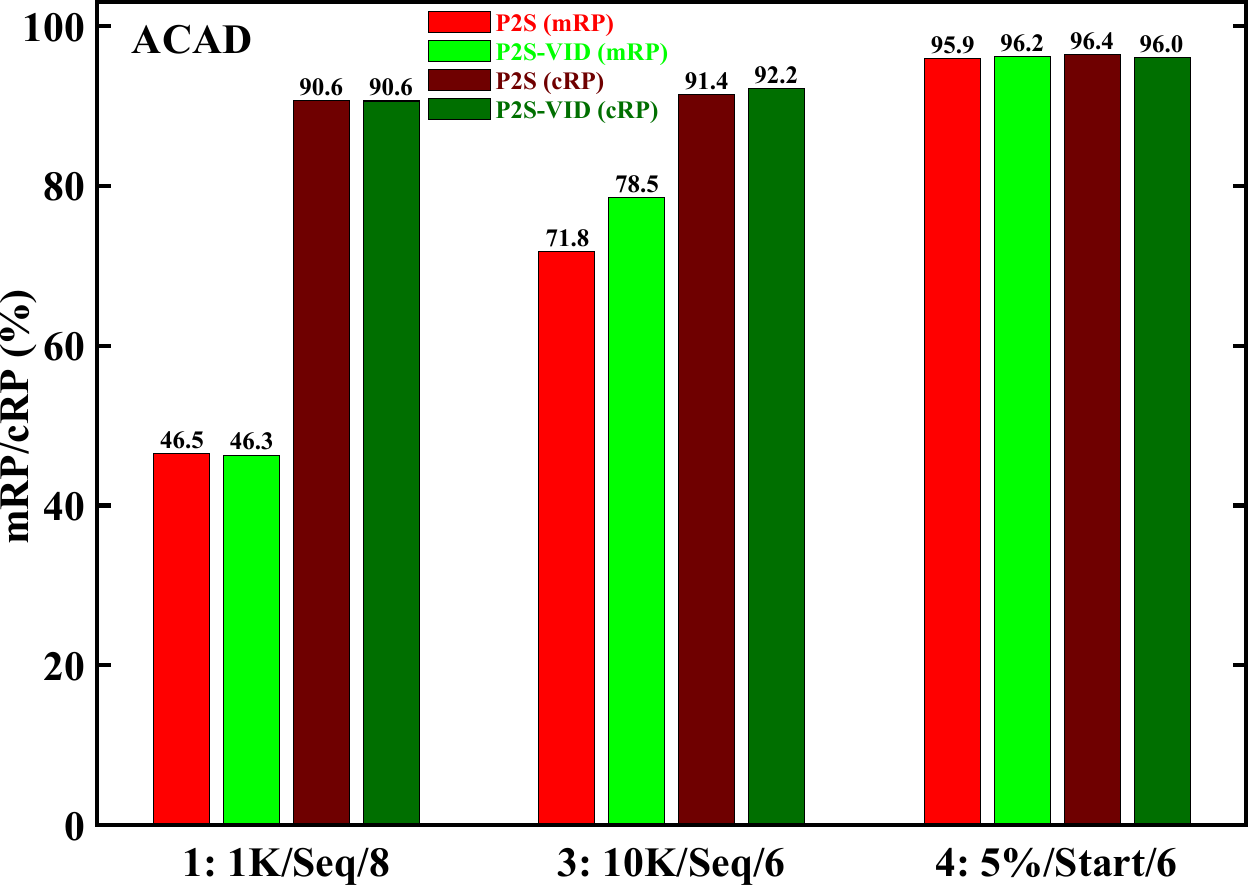}
	\caption{		
		Results on ACAD configurations $\#1$, $\#3$ and $\#4$ \cite[Table~4]{Singh2020_Animal_det} in terms of both mRP and cRP \cite[Sec.~4.1]{Singh2020_Animal_det} shown respectively with lighter and darker shades of red and green.		
		P2S and P2S-VID respectively refer to the static and video detection Pix2Seq models.
		P2S-VID uses middle fusion architecture with $N=2$ and was trained with frozen backbone and class token weight equalization.
	}
	\label{fig_acamp}
\end{figure}
This section presents results comparing our proposed token-based
video detection model with the baseline Pix2Seq static detection model \cite{p2s} along with some state of the art video detection models.
Unless otherwise specified, all Pix2Seq models have the $640\times 640$ ResNet-50 as their backbone and the video models use the middle-fusion video architecture (Sec. \ref{middle_fusion}) with $N=2$.
For the sake of brevity, Pix2Seq static and video detection models are referred to as \textbf{P2S} and \textbf{P2S-VID} respectively for the remainder of this paper.
We experimented with many different configurations or variants of P2S and P2S-VID (\ref{exp}) but this section only summarizes the best results we found.
The specific model configurations that we have included here for each dataset are detailed in Table \ref{tab:model_configs}.
Note that we were only able to train a small fraction of all the models we would have liked to have trained due to limited time and computational resources, so these results very likely do not indicate the best performance that these models are capable of, especially in the case of the video models with larger values of $N$.

\subsubsection{Summary}
\label{res_overview}
Following are the key takeaways from the results presented in the remainder of this paper:
\begin{itemize}[left=0pt,label=\textendash]
	\item P2S-VID models exhibit strong signs of being bottlenecked, especially for larger values of $N$, by the low batch sizes (\ref{exp_batch}) and possibly also the relatively small amounts of training data we had to use.
	\item Pix2Seq models are better at localizing objects than classifying them correctly so that the class-agnostic performance of P2S-VID tends to compare more favourably with P2S than its overall performance.
	\item Related to the last point is that Pix2Seq models are relatively less robust to class imbalance and tend to overfit to the more numerous class.
	\item This problem of poor classification performance can be partially ameliorated by equalizing the weights assigned to class tokens during training so that each class token has the same weight as all of the corresponding bounding box coordinate tokens combined (\ref{exp_cls_eq}).
	\item P2S-VID performs appreciably better overall than the baseline P2S but this improvement is mainly due to the output redundancy (Sec. \ref{Temporal_Windows}) obtained by using $T<N$ and this performance advantage mostly disappears by setting $T=N$.
	\item There is no consistent improvement in performance with increase in $N$  (\ref{exp_vid_len}).
	\item Static models trained to predict video outputs by processing only the first frame in each video temporal window  (\ref{exp_arch_static}) are able to keep up with the video models surprisingly well, even for large values of $N$, indicating that the latter are not able to make sufficient use of the video information.
\end{itemize}

\begin{figure*}[!t]
	\centering
		\includegraphics[width=0.48\textwidth]{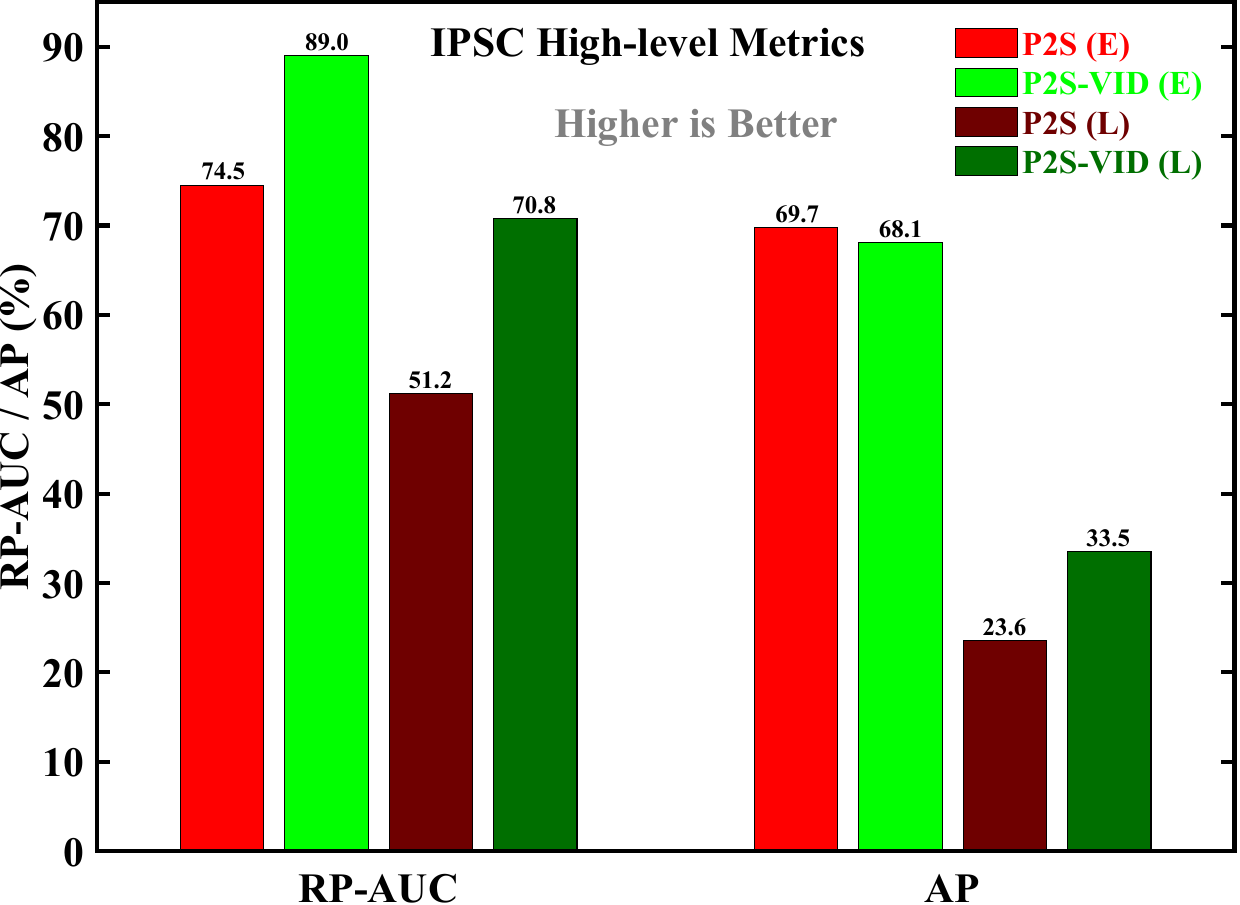}
		\includegraphics[width=0.50\textwidth]{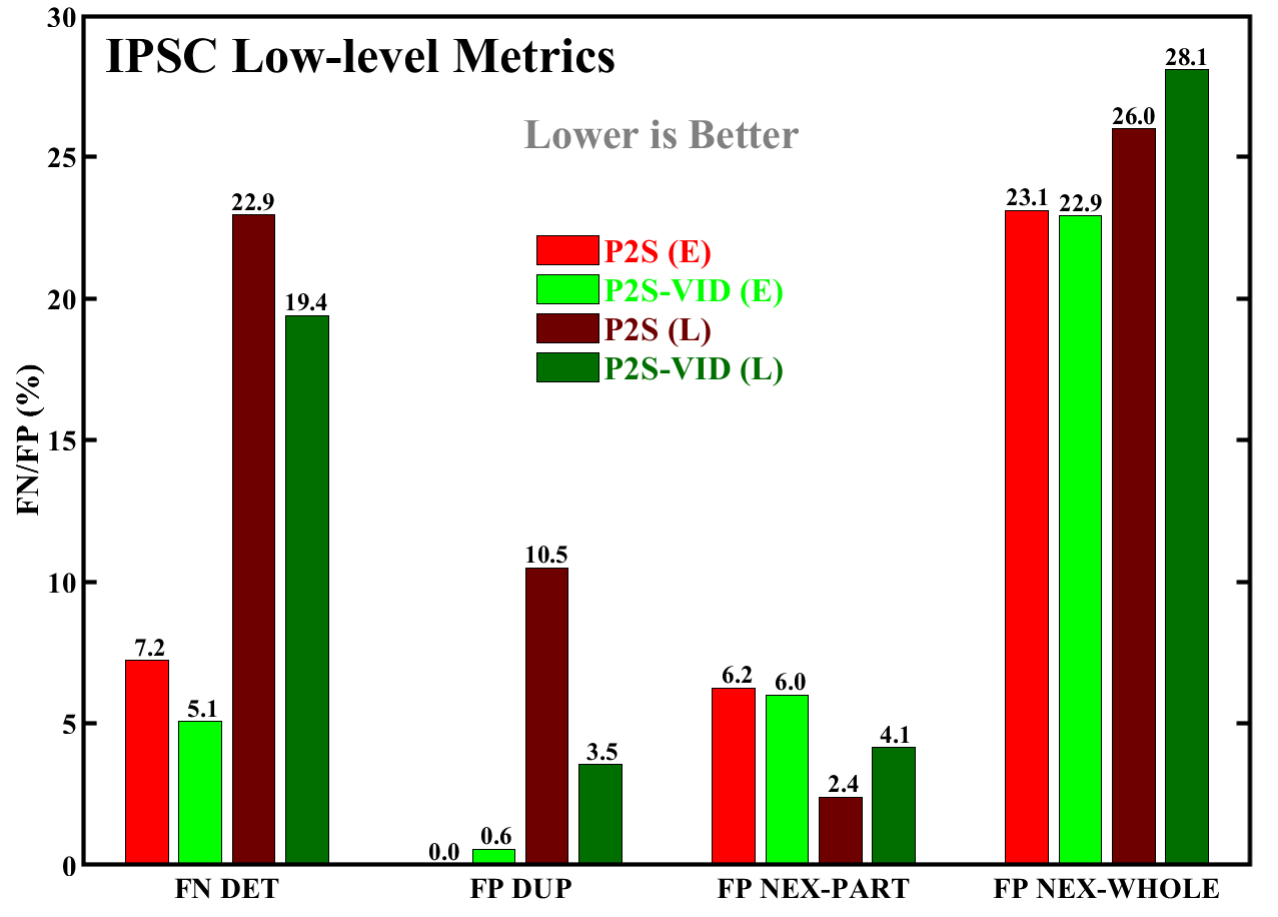}
	\caption{
		Comparing P2S and P2S-VID on IPSC dataset in terms of (left) high-level and (right) low-level detection metrics.
		Results are provided for both early and late-stage training configuration, shown respectively with lighter and darker shades of red and green and denoted in the legend with the suffixes E and L.
	}
	\label{fig_ipsc_det_high_low}
\end{figure*}
%

\subsubsection{ACAD}
\label{vid_det_acad}
Fig. \ref{fig_acamp} summarizes performance on all three configurations of the ACAD dataset.
Note that, even though config \#4 has the smallest training set, it is the easiest to handle from a classification standpoint because the training set contains frames from nearly every  sequence in the test set and therefore the model is able to train on every combination of backgrounds and foregrounds available therein.
While the other two configs contain more frames overall, they have no overlap between the training and test sequences and offer the models access to far fewer combinations of foregrounds and backgrounds in the test set.
P2S-VID mostly performs about the same as P2S, but it does significantly outperform the latter in the challenging config \#3 which is the only one that has enough frames to at least partially alleviate the bottleneck imposed by the relatively small datasets.

\subsubsection{IPSC}
\label{vid_det_ipsc}

Fig. \ref{fig_ipsc_det_high_low} shows the results for both configurations of this dataset in terms of both high-level and low-level detection metrics.
P2S-VID significantly outperforms P2S on nearly every metric especially on the much more challenging late-stage configuration.
The only minor exception is FP-NEX which can be attributed to the redundancy involved in combining detections from multiple temporal windows.
As mentioned in \cite[Sec.~3.4.2]{Singh23_ipsc}, FP-NEX essentially measures how many of the unlabeled cells are incorrectly detected and classified as IPSCs, something which the relatively poor classification ability of the language models makes them susceptible to.

\subsubsection{UA-DETRAC}
\label{vid_det_detrac}

Fig. \ref{vid_len} shows a summary of results on the UA-DETRAC dataset for P2S and P2S-VID with $N=2$ to $N=32$.
We had hoped that the size of this dataset would allow the severe bottleneck on the video models to be at least partially overcome but that turned out not to be the case.
All the models, including P2S, appear to be limited by $B$ here, judging by the identical ceiling of around $85\%$ that they all reach before $N$ becomes too large.
As seen in the example of ACAD \#3 (Fig. \ref{batch_acad}), $B$ seems to be an even more important bottleneck than the size of the dataset.
In fact, the optimal $B$ probably increases with the size and complexity of the dataset, so the models are likely to be even more bottlenecked on UA-DETRAC than ACAD.
It is true that the ACAD example demonstrated performance limitation only on the classification task and this is not relevant with UA-DETRAC since it has only one class.
However, UA-DETRAC has much longer and more complex trajectories than ACAD.
This makes its localization task a lot more challenging and therefore just as likely to be bottlenecked as classification on ACAD.
P2S-VID models upto $N=8$ are able to mostly match P2S with $T=1$ but the performance drops sharply after that, thus confirming the inadequacy of our existing hardware for training such large models.
We would expect that all the models, including P2S, would be able to reach $> 90\%$ if trained with large enough B.

We would like to conclude this section with a recent result which is a promising step in this direction.
We generated this by training models on two Tesla A100 $80$ GB GPUs that we rented to produce publication-quality results that are competitive with the current state of the art.
VSTAM \cite{fujitake2022_VSTAM} is the top-ranked model on the leaderboard \cite{detrac_leaderboard} at the time of this writing and has been so since it was released nearly 3 years ago.
However, as shown in Table \ref{tab:detrac_new}, one of our P2S-VID models was able to outperform VSTAM by $0.75\%$.
This model uses the late-fusion architecture which turns out to significantly outperform the other two fusion schemes once the batch size bottleneck is alleviated, as we had hypothesized in Sec. \ref{late_fusion} based on its theoretical advantage of cross-attending every token with each of the $N$ frames.
Also, this model was trained with the smallest possible video length of $N=2$ because even $160$ GB GPU RAM is apparently insufficient to relieve the bottleneck on models with higher $N$ (Table \ref{tab:detrac_new_detailed}).
We were was hoping to rent 4 or even 8 of these GPUs to better exploit the video length but these configurations were unfortunately not available.
This result is perhaps the most convincing evidence we have that most of the P2S-VID results in this paper are of bottlenecked models and do not represent their true potential.

\begin{table}[t]
	\centering
	\caption{
		Comparing P2S and P2S-VID with the current state of the art video detection models on UA-DETRAC.		
	}
	\begin{tabular}{|c|c|}
		\hline		
		\textbf{Model}                          & \textbf{mAP (\%)}                     \\
		\hline
		TFEN \cite{fujitake2021_TEFN}                                   & 82.42                                 \\
		YOLOv3-SPP \cite{kim2018_YOLOv3_SPP}                              & 84.96                                 \\
		MSVD\_SPP \cite{kim2019_MSVD_SPP}                              & 85.29                                 \\
		SpotNet \cite{perreault2020_spotnet}                                 & 86.8                                  \\
		FFAVOD-SpotNet \cite{perreault2021_ffavod}                         & 88.1                                  \\
		\rowcolor[HTML]{FFF2CC} 
		VSTAM \cite{fujitake2022_VSTAM}                                  & 90.39                                 \\
		P2S                        & 88.62                                  \\
		\rowcolor[HTML]{E2EFDA} {\color[HTML]{000000} \textbf{P2S-VID}} & {\color[HTML]{000000} \textbf{91.14}}\\
		\hline
	\end{tabular}
	\label{tab:detrac_new}
\end{table}

\begin{figure*}[!t]
	\centering
	\includegraphics[width=0.495\textwidth]{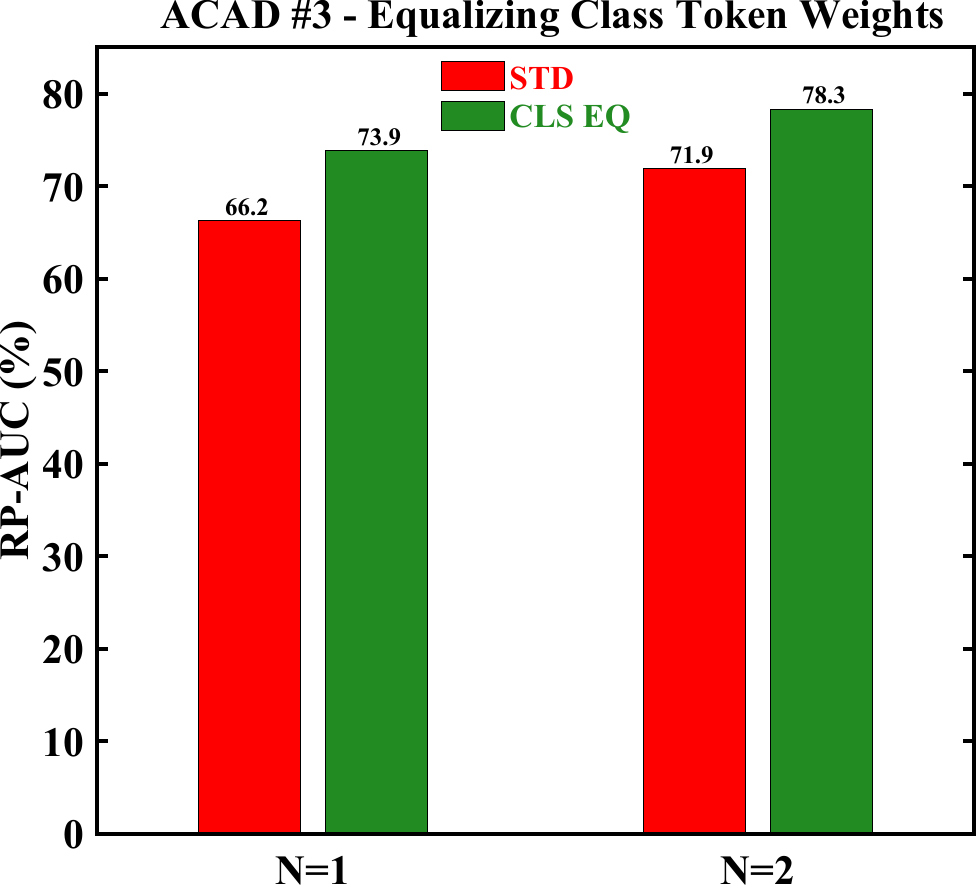}
	\includegraphics[width=0.495\textwidth]{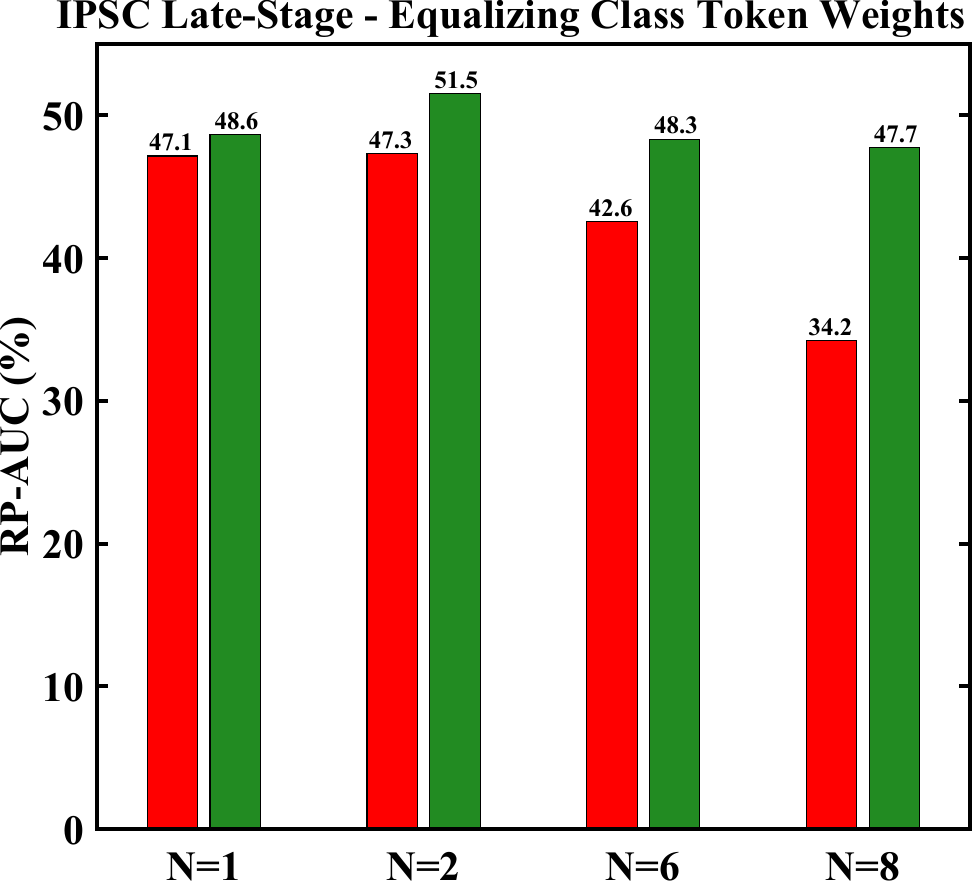}
	\caption{
		Performance impact of equalizing the class token weights on (left) ACAD $\#3$ and (right) IPSC late-stage datasets.	
		\textit{N=1} denotes the baseline P2S model.
		Models trained with and without class token weight equalization are respectively shown in green and red and denoted with \textit{CLS EQ} and \textit{STD} in the legend.
		Note the different Y-axis limits in the two plots.
	}
	\label{cls_eq}
\end{figure*}

\section{Ablation Study}
\label{exp}
This section presents the results of our experimentation with some of the important model parameters that were useful in finding the optimal models that are reported in Sec. \ref{res_overview}.
We used the IPSC late-stage dataset for most of these experiments since it is small enough to allow training a large number of models while at the same time also being challenging enough to be able to discriminate between these models.

\subsection{Equalizing Class Token Weights}
\label{exp_cls_eq}
A possible reason for the poor classification performance of Pix2Seq models, especially in object detection, might be that the weight assigned to the class tokens during training is equal to that assigned to every single coordinate token.
Since the number of coordinate tokens is $4\times N$ times greater than the number of class tokens, this effectively means that the localization task is assigned $4\times N$ times greater weightage than the classification task.
In order to address this disparity, we trained detection models with each class token assigned the same weight as all of the corresponding coordinate tokens combined.

As shown in Fig. \ref{cls_eq}, this does improve the classification performance appreciably, particularly on large datasets and with higher vlues of $N$.
The degree of improvement also increases with $N$, especially in the IPSC case.
This makes sense since the factor by which coordinate tokens are over-weighted with respect to class tokens without equalization increases linearly with $N$ as does the overemphasis on the localization task.
This improvement does come at the cost of significantly slower training, at least in the case of larger datasets.
For example, ACAD \#3 took 700K iterations and close to 12 days to reach convergence with class equalization and only 300K iterations and just over 5 days without it.
These disparities were not as strongly marked on the IPSC dataset, where both sets of models took approximately the same amount of time to converge.

\subsection{Training Batch Size (B)}
\label{exp_batch}
\begin{figure*}[!t]
	\centering
	\includegraphics[width=0.495\textwidth]{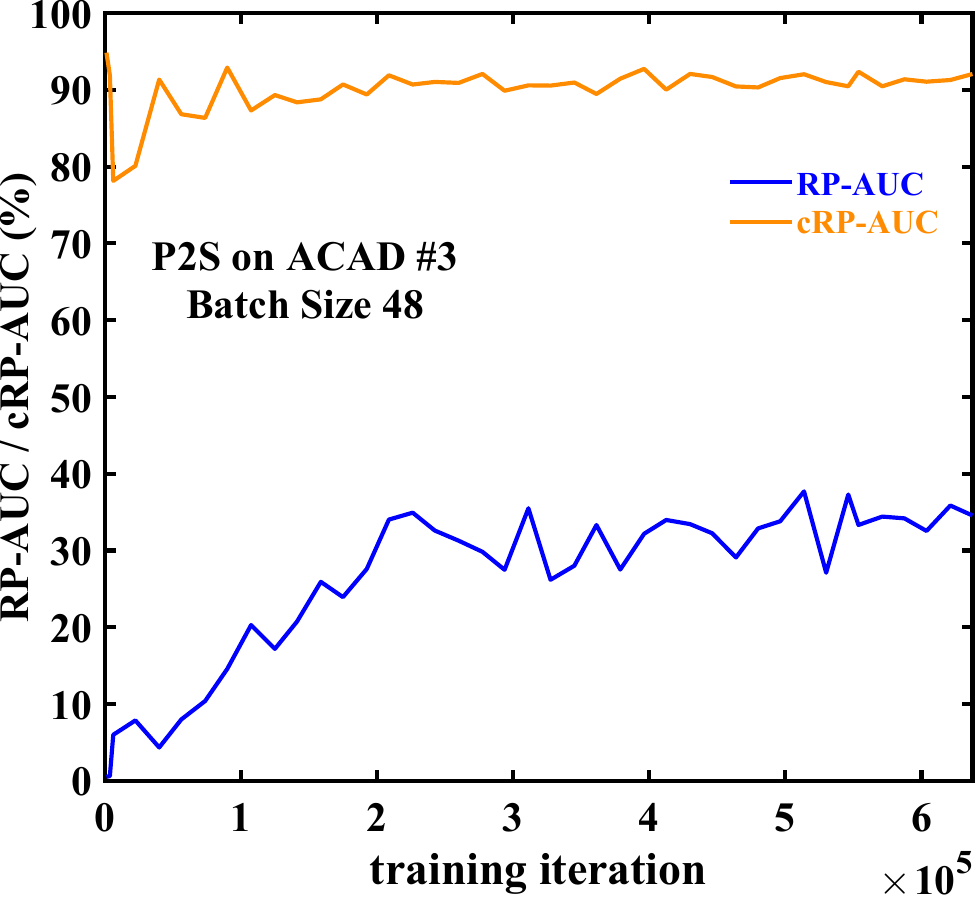}
	\includegraphics[width=0.495\textwidth]{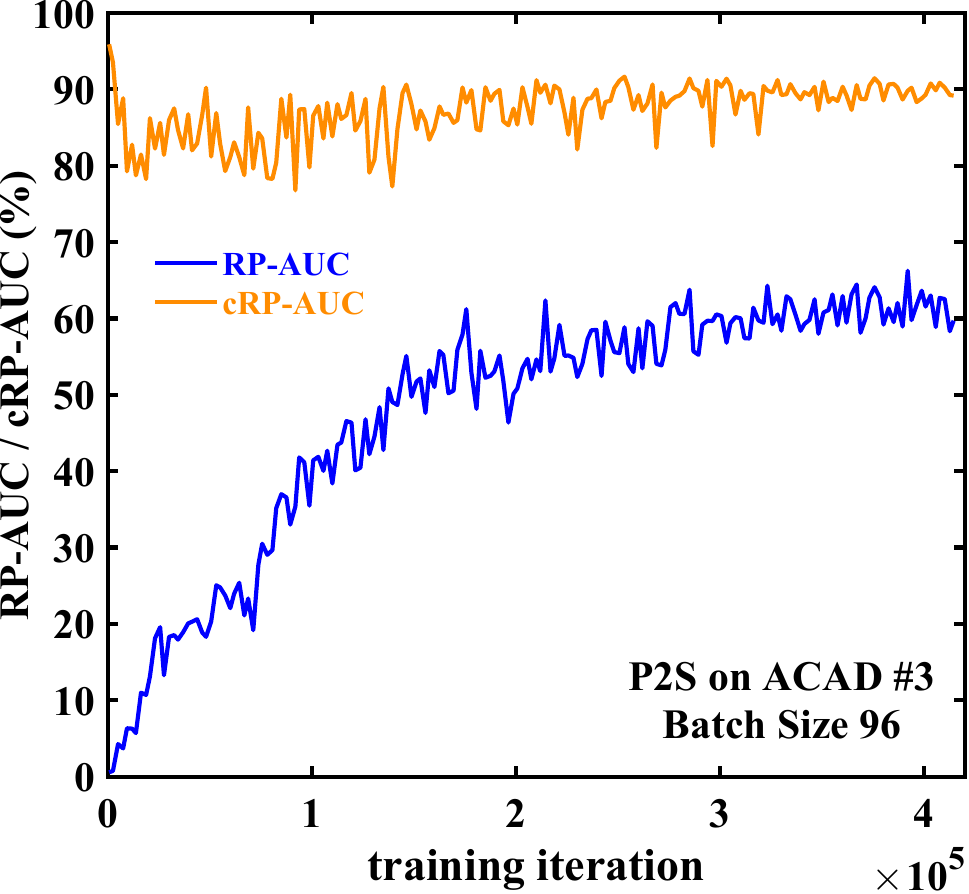}
	\includegraphics[width=0.495\textwidth]{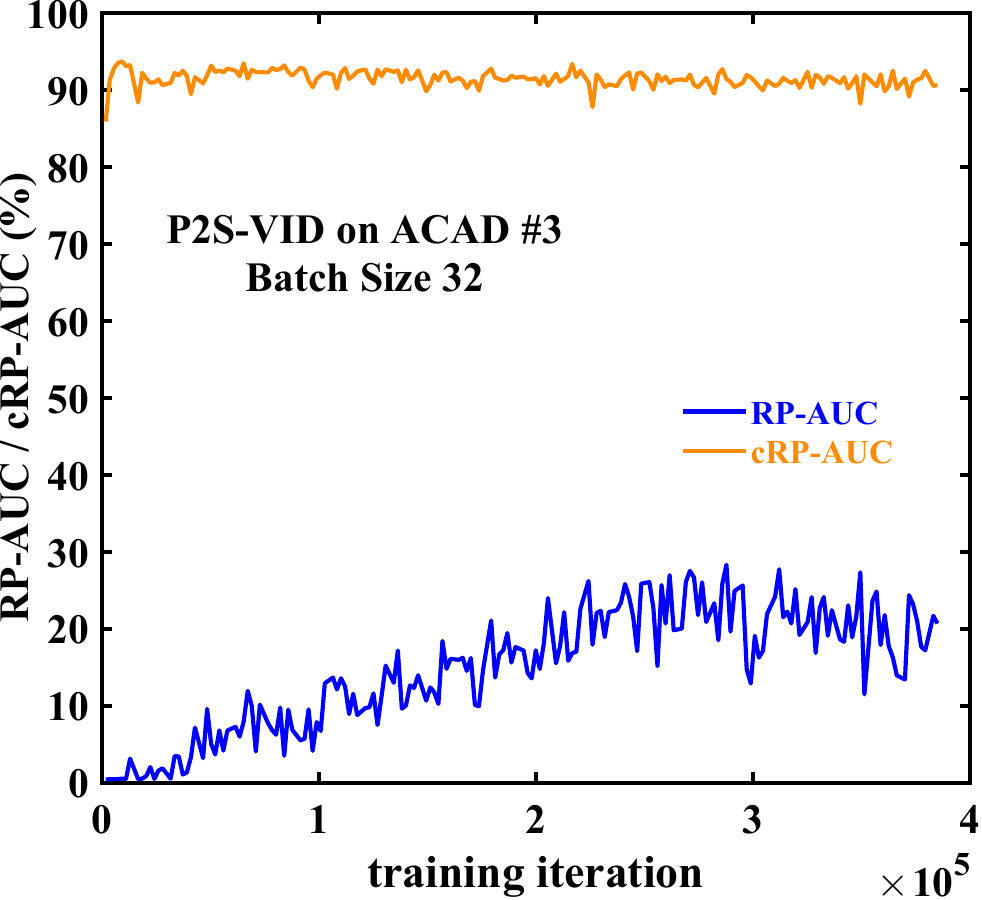}
	\includegraphics[width=0.495\textwidth]{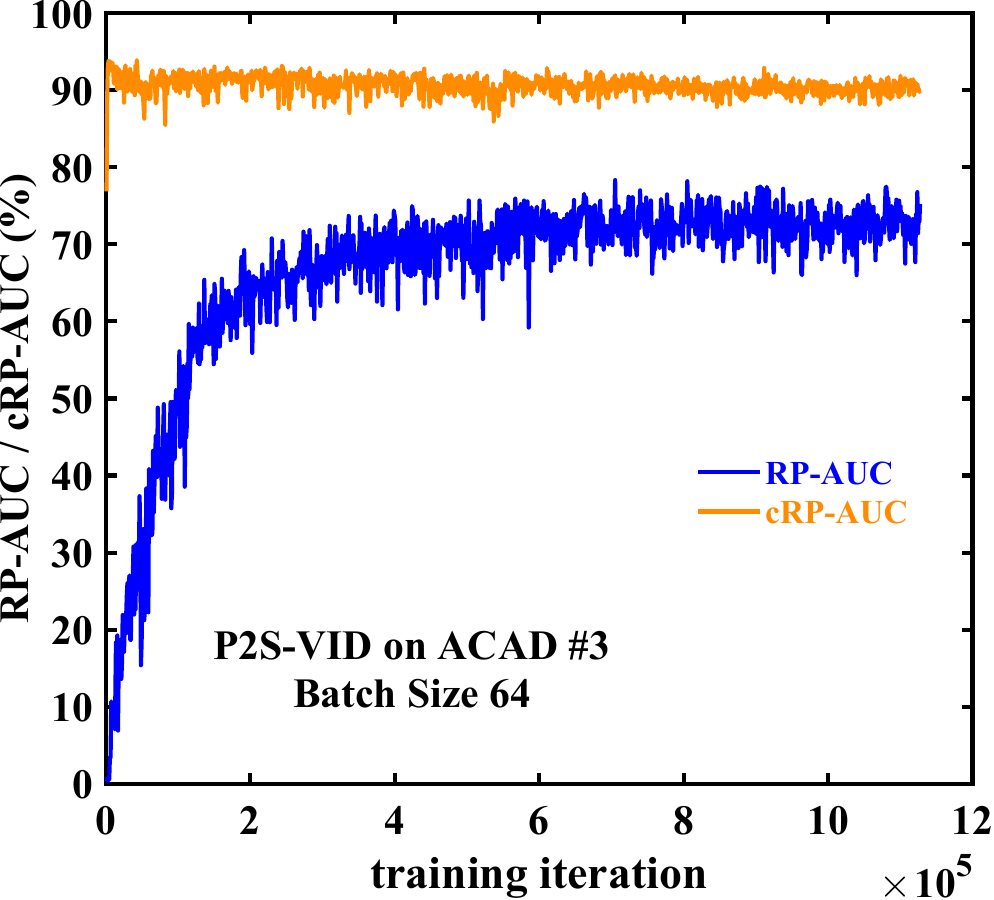}
	\caption{
		Impact of batch size on validation performance when training (top) P2S and (bottom) P2S-VID on ACAD \#3.
		The left plots show the results for training on a single RTX 3090 GPU with batch sizes 48 and 32 while the ones on the right show dual-GPU training with batch sizes 96 and 64.
	}
	\label{batch_acad}
\end{figure*}
Extensive experiments have shown to us that the training batch size matters a lot more for large datasets like ACAD and UA-DETRAC rather than smaller ones like IPSC.
It also has a greater impact on video models than static ones, especially with larger values of $N$.
Finally, it has far greater impact on the classification task (i.e. RP-AUC) than localization (i.e. cRP-AUC).
Fig. \ref{batch_acad} shows an example for both P2S and P2S-VID training on ACAD \#3.
It can be seen that the peak RP-AUC on the validation set nearly doubles from $37\%$ to $66\%$ for P2S and nearly triples from $28\%$ to $78\%$ for P2S-VID.
On the other hand, cRP-AUC peaks at just above $90\%$ in all four cases and this is reached in just a few thousand iterations as opposed to RP-AUC which takes from 200K to 700K iterations to attain its plateau.
The significantly greater relative increase in the case of P2S-VID (2.76 times vs. 1.75 times for P2S), along with the much lower absolute peak with the smaller batch size ($28\%$ vs. $37\%$) confirms the much greater bottleneck faced by the video models.

P2S-VID here uses only $N=2$ and this bottleneck only gets worse as $N$ increases.
In our experience, and subject to the size and complexity of the dataset, the optimal batch size required for a video model to achieve its full potential increases at least linearly with $N$.
However, the GPU memory required to train a video model also increases linearly with $N$ even if the batch size remains unchanged.
Since the total amount of available GPU memory is fixed, we have to actually \textit{decrease} the batch size linearly with $N$.

\subsection{Video Architecture}
\label{exp_arch}
\begin{figure*}[!t]
	\centering
	\includegraphics[width=0.495\textwidth]{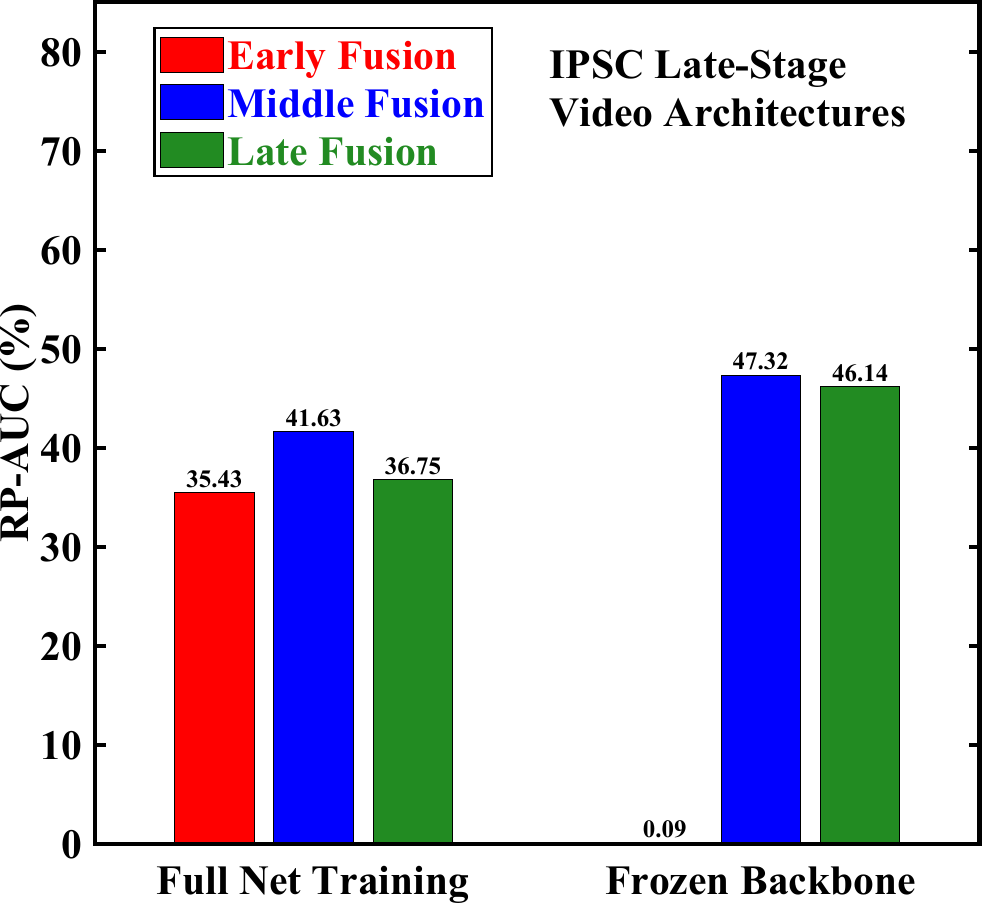}
	\includegraphics[width=0.495\textwidth]{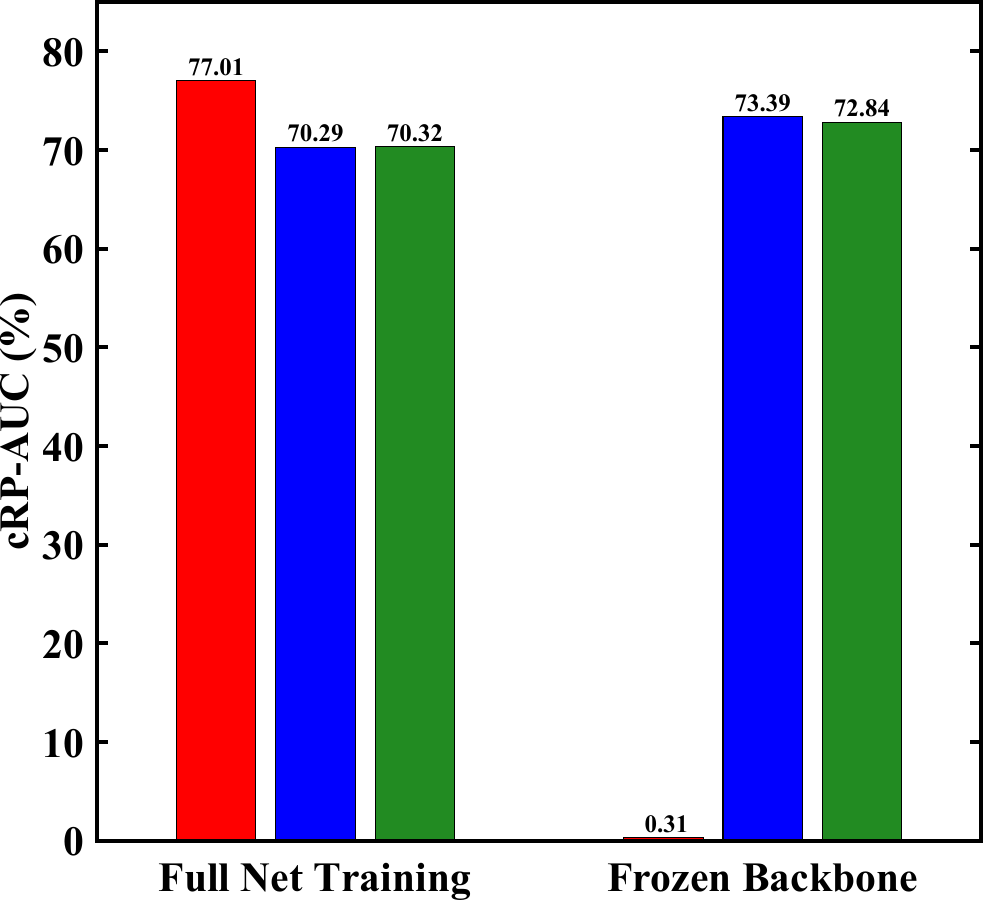}
	\caption{
		Comparing the three video architectures (Sec. \ref{vid_net_arch}) with $N=2$ and trained with and without frozen backbone.
		The left and right plots show RP-AUC and cRP-AUC respectively.
		All models were trained on the IPSC late-stage dataset. 		
	}
	\label{vid_arch}
\end{figure*}
Fig. \ref{vid_arch} shows comparative results for the three video architectures (Sec. \ref{vid_net_arch}) with $N=2$.
Middle-fusion outperforms the other two models in terms of mRP while early-fusion performs best in terms of cRP.
We have seen this trend of early-fusion models performing much better on the localization task than classification in other models we have trained as well.
The video Swin transformer backbone we are using for early-fusion was pre-trained for human action-recognition and therefore unsurprisingly finds it easier to localize objects than to classify them correctly.
Although late-fusion seems slightly inferior to middle-fusion in these results, we have seen it outperform the latter with similar margins on other values of $N$ so, on the whole, the two models can be said to perform at par.

\subsubsection{Frozen Backbone}
\label{exp_fbb}
We would have expected that the entire Pix2Seq network would need to be retrained in order to work well on a new task – either video detection or semantic segmentation - but this turned out not to be the case.
As shown in Fig. \ref{vid_arch}, keeping the backbone frozen actually leads to significantly better video detection performance for both middle and late-fusion architectures.
However, this is not true for early-fusion, which completely fails to learn anything useful with its video Swin transformer backbone frozen.
This backbone was pretrained for action recognition with conventional modeling and it seems that weights optimized for conventional modeling cannot generalize to language modeling without retraining.
Nevertheless, these weights are useful in fine-tuning the network for language modeling since we found that training the early-fusion models without loading the pretrained weights for its backbone leads to significantly worse performance.
Note that freezing the backbone allows us to use much higher batch sizes (e.g. $80$ vs. $20$ with $N=2$) which must partially account for this improvement as well.

\subsubsection{Video Output with Static Input}
\label{exp_arch_static}
\begin{figure*}[!t]
	\centering
	\includegraphics[width=0.49\textwidth]{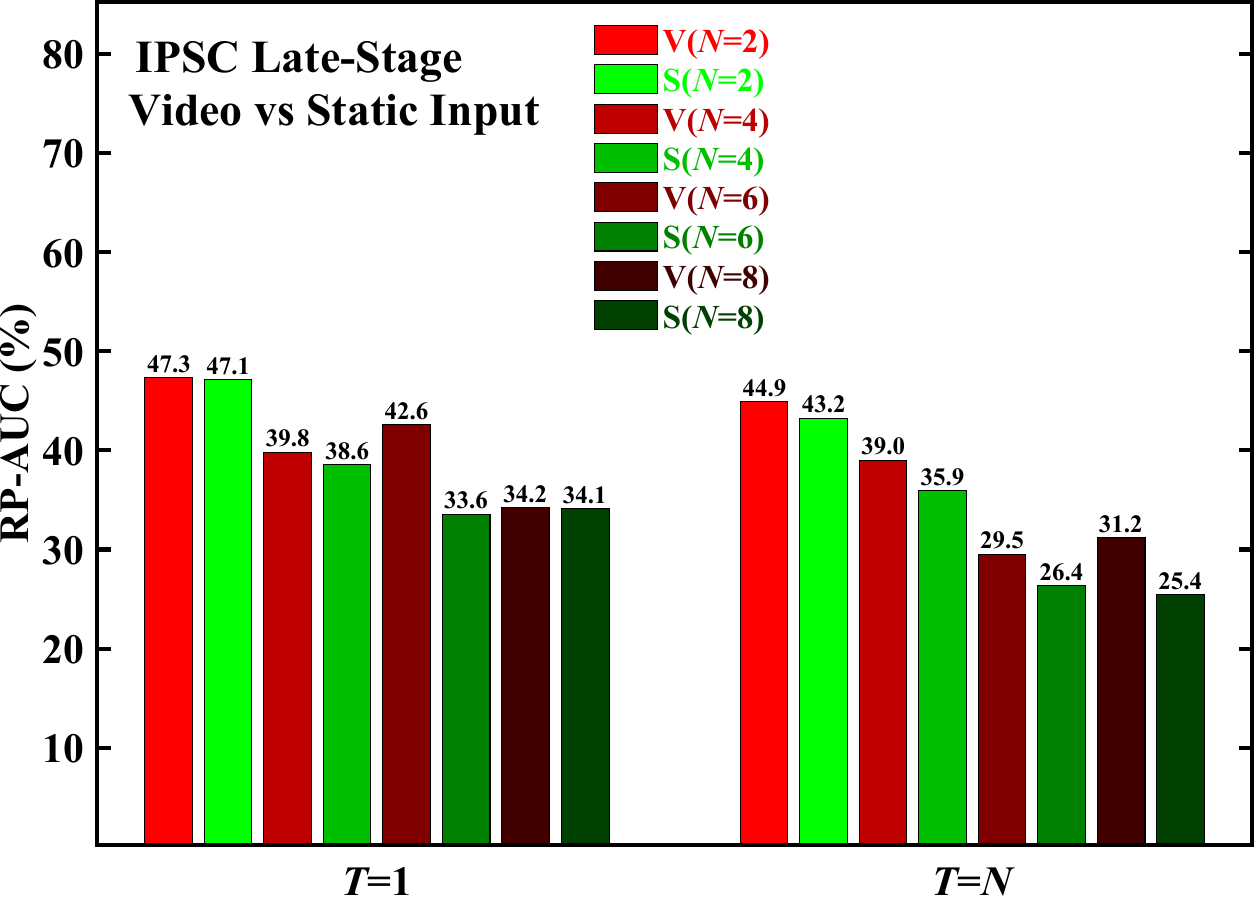}
	\includegraphics[width=0.49\textwidth]{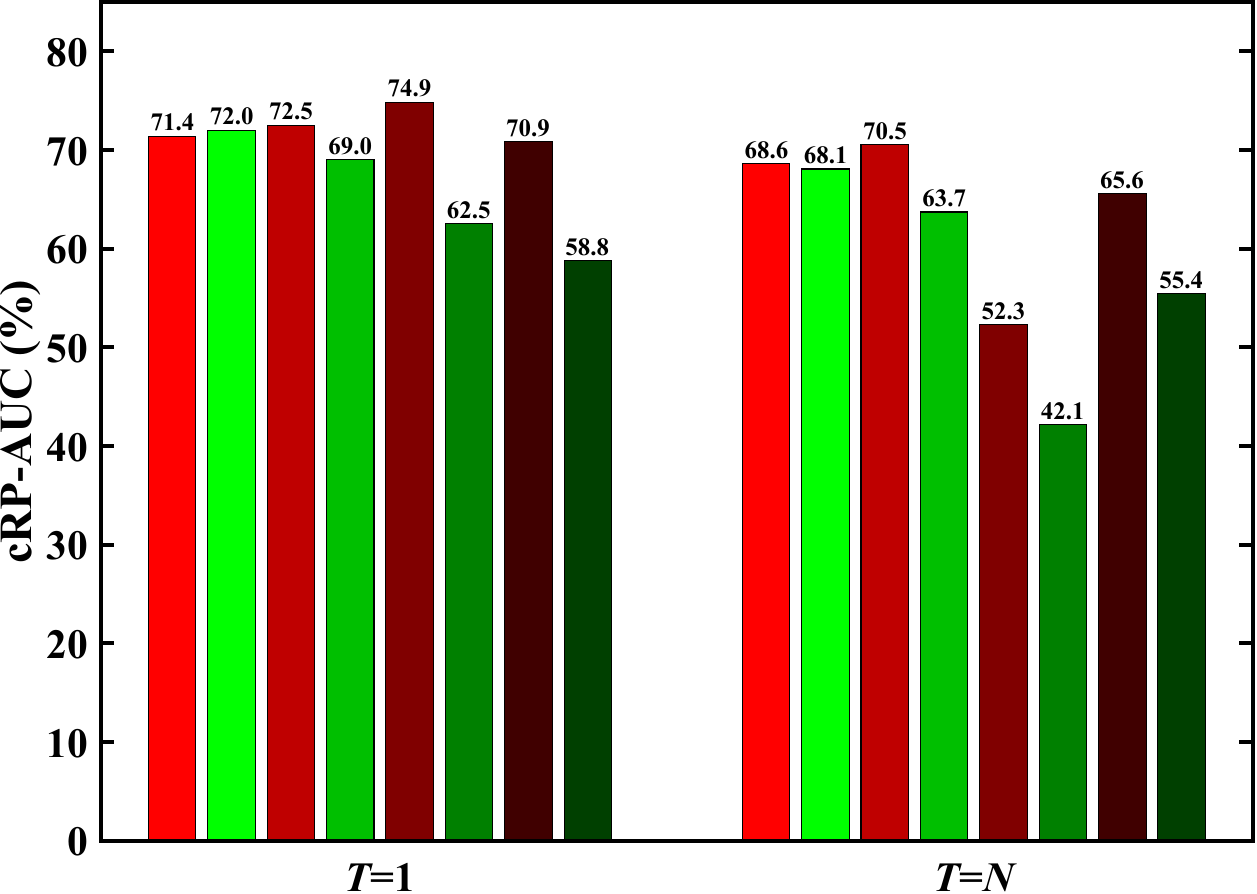}
	\caption{		
		Performance impact of replacing $N$ video frames with only the first frame in the sequence as input to P2S-VID models.
		The two cases are respectively shown in shades of red and green and denoted with \textit{V} and \textit{S} in the legend.
		Darker shades represent higher $N$.
		All video input models used the middle-fusion architecture without class token weight equalization.
		The left and and right plots show RP-AUC and cRP-AUC respectively.
		All models were trained on the IPSC late-stage dataset. 		
	}
	\label{static_vid_det}
\end{figure*}
We wanted to find out how much useful information the network is able to learn from video frames so we trained static models to predict video output (either detection or segmentation) using only the first frame $F_1$ in each video temporal window as input.
This means that the model is trained to produce the same output as a P2S-VID or P2S-VIDSEG model but it only has access to the first frame in each temporal window,  rather than all $N$ frames.
It therefore needs to use the first frame to predict the contents of the future $N-1$ frames in the sequence.
Since there is no more video input, we can use the baseline P2S architecture for these models.
We trained models with $N=2$, $N=4$, $N=6$ and $N=8$, all with the backbone frozen, and all on the IPSC late-stage dataset.
Note that these static-video models can be trained with the same (and much larger) batch size as P2S, irrespective of $N$.
This gives them an advantage over the true video models whose batch size decreases linearly with $N$.
The video stride $T$ is important in evaluating these models since $T=1$ ensures that each frame would be the first frame in some temporal window so that the static input models can output valid boxes only for the first frame and still not be penalized during inference.
This is unlikely to happen in practice since the model is trained to output boxes for all $N$ frames and therefore will be penalized during training for learning a simple strategy like this.

Results are shown in Fig. \ref{static_vid_det}.
Static input models show surprisingly little performance loss over the video models even for $N$ as high as 8, though this loss is greater for $T=N$ than $T=1$ as expected.
Similarly, the performance loss predictably increases with $N$, except for the odd case of $N=8$ with $T=1$ (Sec. \ref{exp_vid_len}).
The performance loss is also greater for cRP-AUC than RP-AUC, which makes sense since the first frame is usually sufficient to classify an object but we need the remaining frames to localize it correctly in those frames.
The fact that static input models are able to perform so well even with $T=N$ is explained at least partly by the fact the cells do not move a lot and this makes the bounding boxes in the IPSC dataset relatively easy to predict.
It is probably also another indicator of the bottleneck on the video models that are unable to make full use of the video information themselves. 
It is also possible, though we consider it unlikely, that the video information fusion schemes we have come up with are not good enough to take full adavantage of this information.

\subsection{Video Length and Stride}
\label{exp_vid_len}
\begin{figure*}[!t]
	\centering
	\includegraphics[width=0.38\textwidth]{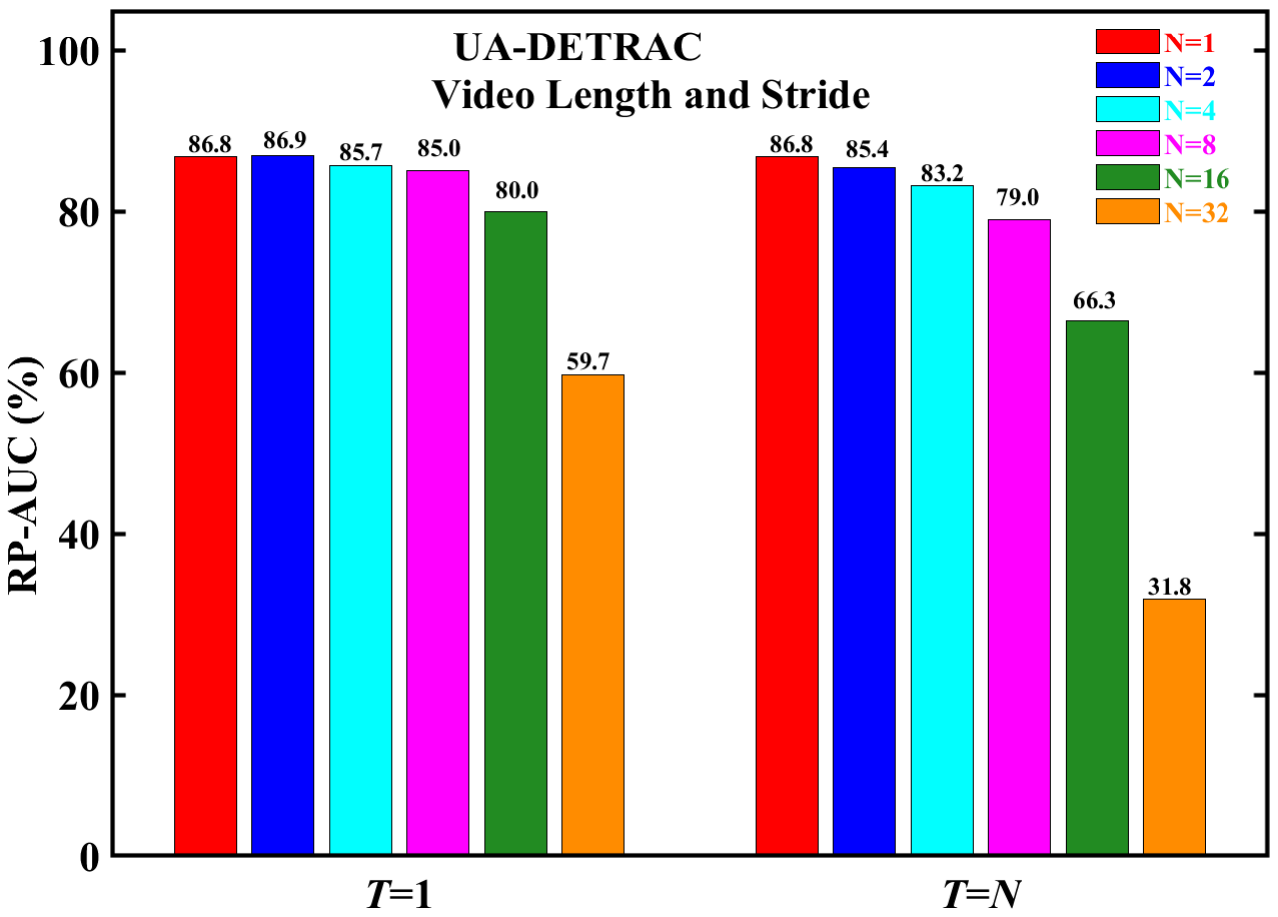}
	\includegraphics[width=0.3\textwidth]{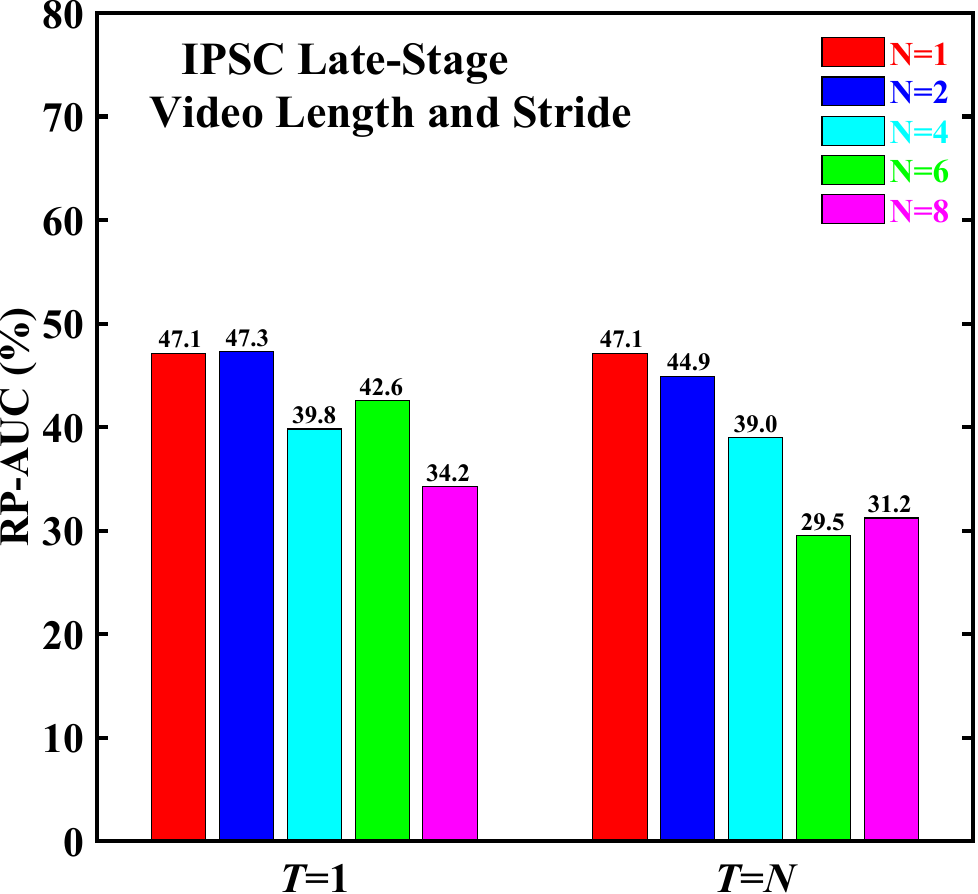}
	\includegraphics[width=0.3\textwidth]{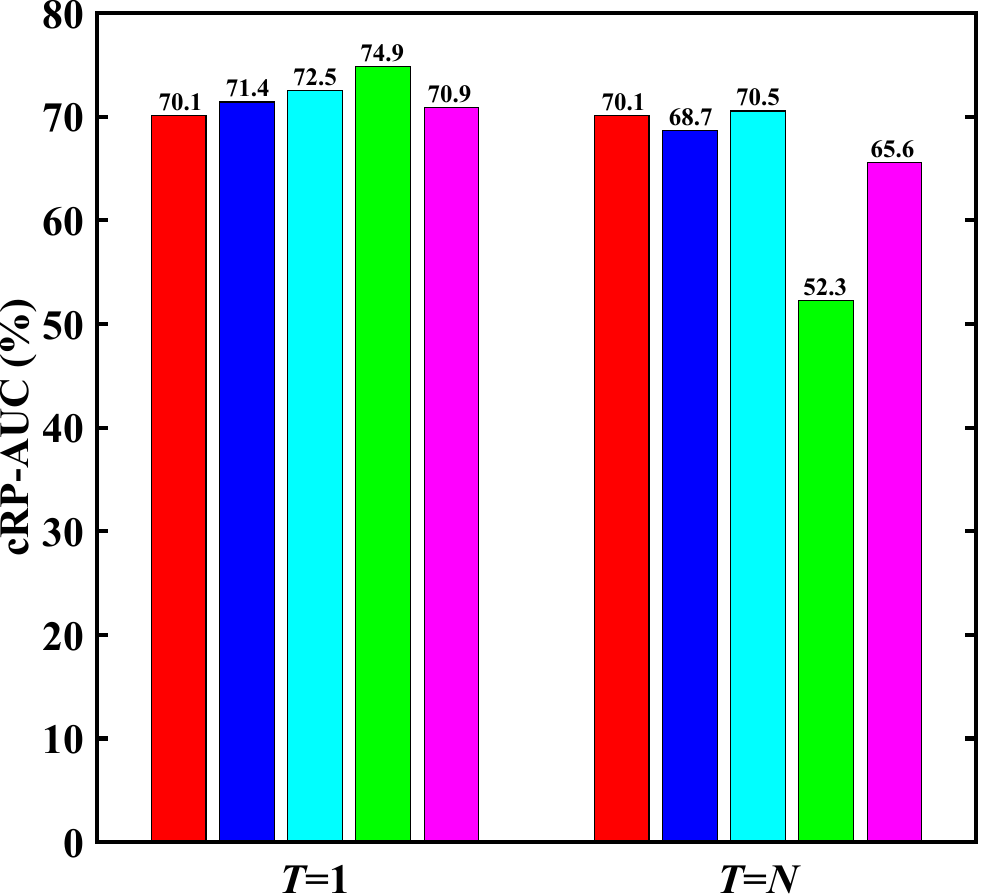}
	\caption{
		Impact of video length $N$ on P2S-VID pereformance over (left) UA-DETRAC and (center and right) IPSC late-stage datasets.
		\textit{N=1} denotes the baseline P2S model and is included for comparison.
	Best viewed under high magnification.		
	}
	\label{vid_len}
\end{figure*}
Fig. \ref{vid_len} summarizes the impact of video length $N$ and stride $T$ on video detection performance over both UA-DETRAC and IPSC late-stage datasets.
Neither dataset shows any consistent improvement in performance with $N$.
Quite the contrary, in fact.
The only case that shows any signs of steady improvement is IPSC cRP-AUC with $T=1$, at least as far as $N=6$.
There is, however, fairly consistent improvement in going from $T=N$ to $T=1$ and the degree of this improvement also increases with $N$, which is to be expected because of the increasing redundancy.

\subsection{1D Coordinate Tokens}
\label{1d}
\begin{figure*}[!t]
	\centering
	\includegraphics[width=0.495\textwidth]{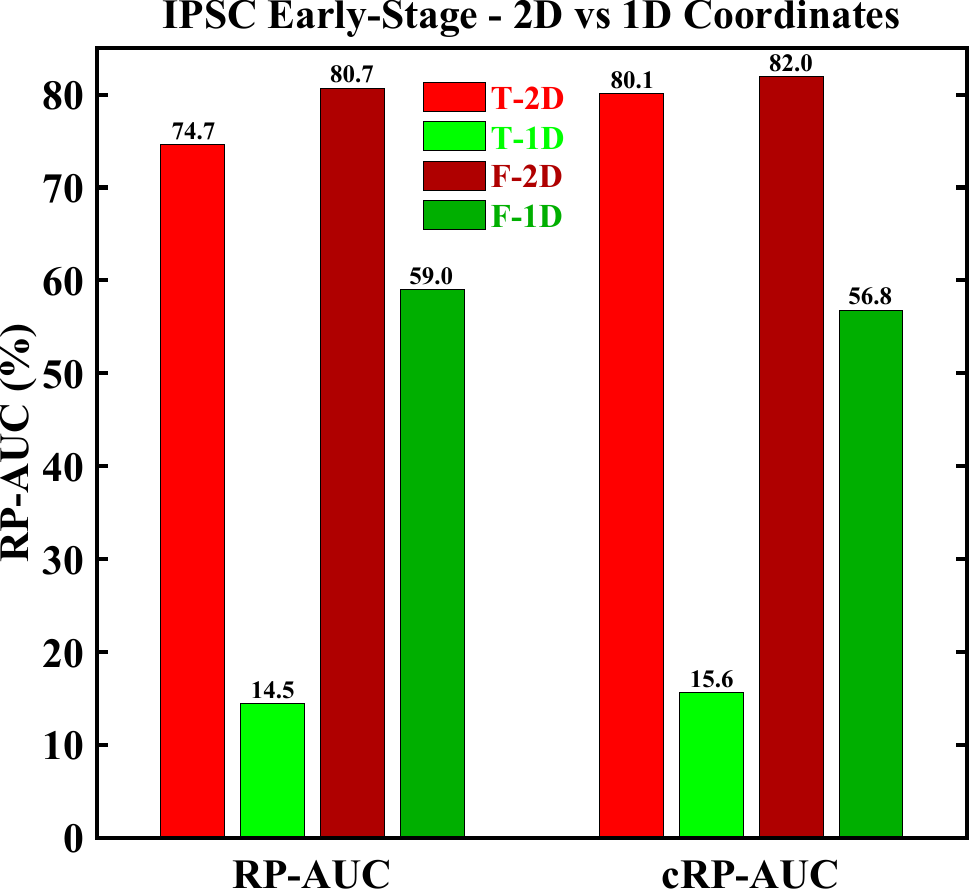}
	\includegraphics[width=0.495\textwidth]{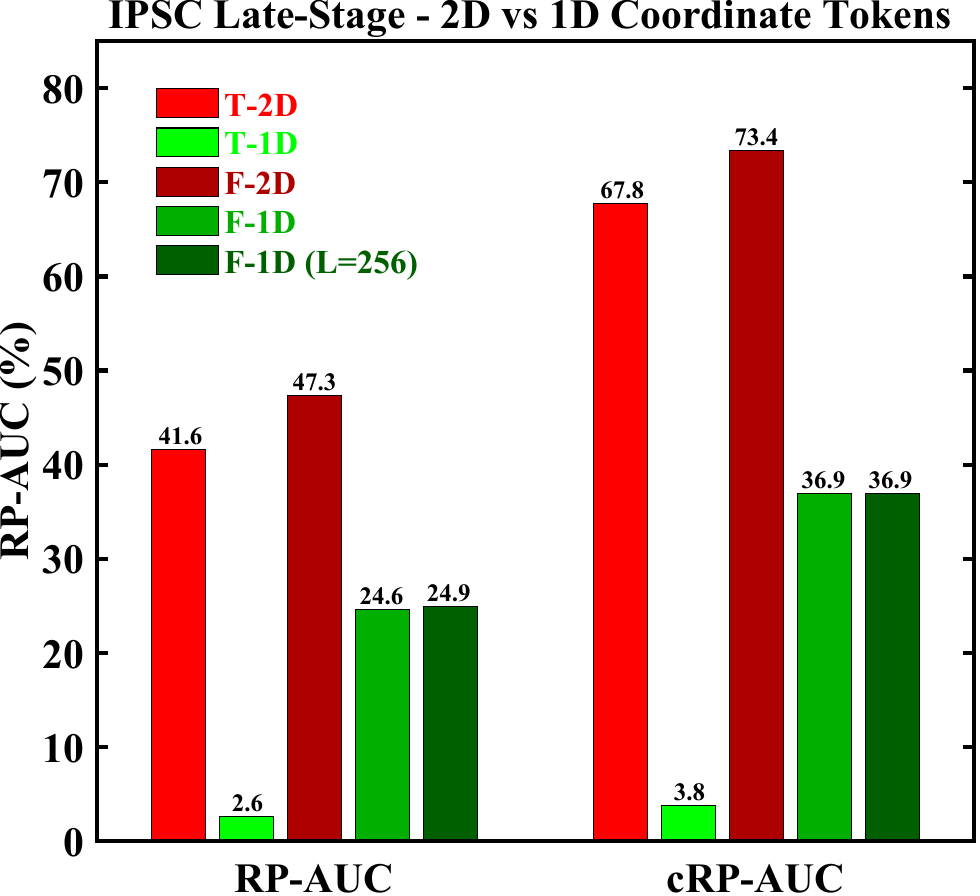}
	\caption{
		Comparing the standard 2D coordinate tokenization of P2S-VID with its 1D variant on the (left) early and (right) late-stage IPSC datasets.
		2D and 1D tokenization models are respectively shown in shades of red and green.
		Models trained with and without frozen backbone are shown respectively in darker and lighter shades and denoted with the prefixes \textit{F} and \textit{T} in the legend. All models use $N=2$.
	}
	\label{1d_ipsc}
\end{figure*}
As mentioned in Sec. \ref{1d_coord}, 1D coordinate tokens can be used to partially solve the problem of $L$ becoming too large as $N$ increases.
We trained a few models to judge the practicability of this approach.
As shown in Fig. \ref{1d_ipsc}, it turned out to be not doable, at least on our existing hardware.
The 1D model was able to achieve barely half the performance of the standard 2D model on the late-stage dataset, although it fared slightly better on the easier early-stage variant. 
An interesting finding here was that training the 1D model without the backbone frozen causes a further sharp decline in performance.
This is surprising since one would expect that learning a new coordinate tokenization  different from the one that was used for pretraining the backbone would benefit from fine-tuning the backbone on the new tokenization, but that is not the case. 
Note that the 2D model was trained with the default Pix2Seq vocabulary parameters $H=2K$ and $V=3K$ while the 1D version had $H=160$ and $V=28K$.
Although both 2D and 1D models were trained with the same $B$, it is possible that this poor performance might be another case of a bottleneck introduced by $B$ since the much higher $V$ in the 1D case probably requires much larger $B$ to work.
This is also supported by the much greater performance advantage of frozen-backbone models with 1D tokenization than with 2D tokenization.
Freezing the backbone allowed the 1D models to be trained with $B=64$ while the non-frozen models were restricted to $B=18$.
The performance drop can also be partially attributed to the decrease in localization accuracy due to the drop in $H$ by a factor of more than 10 from $H=2000$ to $H=160$.
We also trained a 1D model with $L$ reduced by half to $256$ to take advantage of the shorter sequences but this turned out to have no impact on the overall performance.

\section{Conclusions}
\label{conclusions}
This paper has introduced a new way to perform object detection in videos by modeling the outputs of these tasks as sequences of discrete tokens.
We have proposed these new methods as another step in the direction of the more general tokenization of visual recognition tasks that has been happening over the last few years through the paradigm of language modeling.
We have presented theoretical arguments for why such tokenization can help to solve the problems consequent upon trying to model the inherently discrete and variable-length outputs that are common in vision tasks with the continuous-valued and fixed-length representations in conventional modeling. 
We have tested these models on a wide range of real-world problems with in-depth experiments to demonstrate their competitiveness with the state of the art in conventional modeling.
Although we have not been able to demonstrate that our method offers significant and consistent performance advantage over conventional models, we do present strong evidence to suggest that this is not due to any intrinsic weakness in the models themselves.
Rather, it is very likely to be a consequence of the bottleneck that is imposed upon these models by the small training batch sizes that we have been constrained to use by our limited computational resources.
Once these constraints are lifted and the models can be trained to their full potential, we are confident that they will be able to justify their theoretical advantages with practical performance benefits.

\appendix

\section{Hierarchical Video Cross-MHA}
\label{hierarchical_cross_mha}
\begin{figure}[!htbp]
	\centering
	\includegraphics[width=0.48\textwidth]{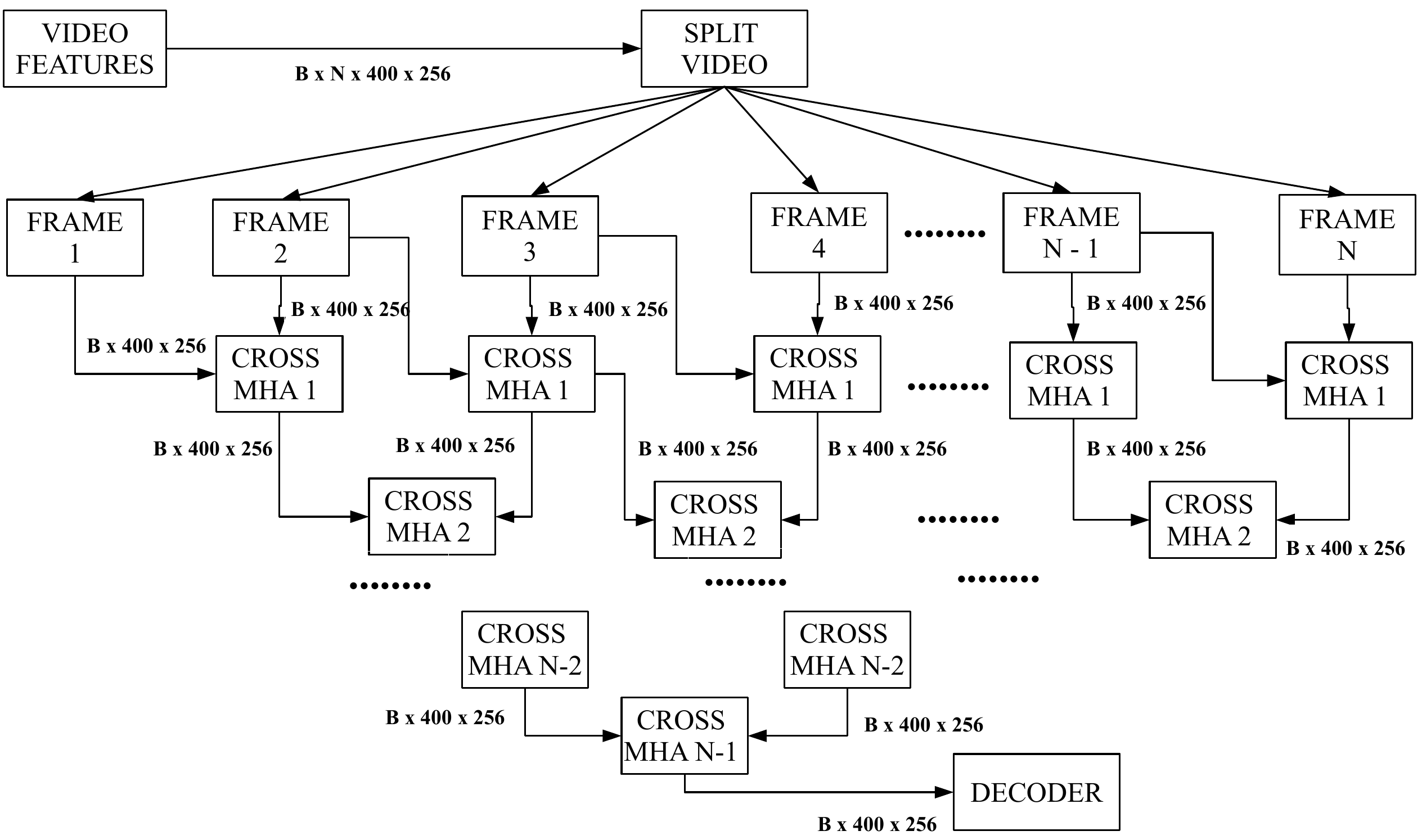}
	\caption{
		Flow diagram for the hierarchical variant of the cross-MHA module in the middle-fusion video encoder.
		Best viewed under high magnification.
	}
	\label{fig:cross_mha_hierarchical}	
\end{figure}
This appendix provides
an alternative way to perform the video cross-MHA operation in the middle-fusion video encoder.
As shown in fig. \ref{fig:cross_mha_hierarchical}, we first apply cross-MHA between pairs of consecutive frames, i.e. ($F_1$, $F_2$), ($F_2$, $F_3$), ($F_3$, $F_4$) and so on to obtain $N-1$ feature maps, each of size $400\times 256$, corresponding to the $N-1$ frame-pairs.
Consecutive pairs from these $N-1$ feature maps are then cross-attended in a second level cross-MHA operation to yield $N-2$ feature maps, again with the same $400\times 256$ size.
This process is repeated for a total of $N-1$ levels of cross-MHA operations to finally yield a single $400\times 256$ feature map that is passed to the decoder.

\section{Model Training}
\label{model_training}
This appendix provides configuration details for the models reported in Sec. \ref{sec_results} as well as the GPU servers used for training these models.

\begin{table}[!htbp]
	\centering
	\caption{
		Performance on UA-DETRAC dataset of P2S and P2S-VID models trained with more GPU RAM to partially alleviate batch size bottleneck.
		Please refer Sec. \ref{vid_det_detrac} and Table \ref{tab:detrac_new} for more details.
	}
	\begin{tabular}{c}
		\includegraphics[width=0.45\textwidth]{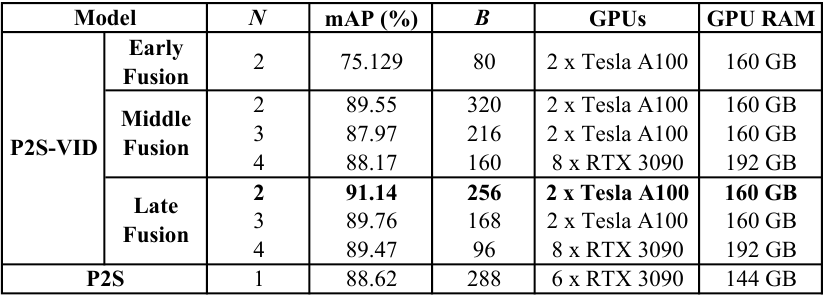}
	\end{tabular}	
	\label{tab:detrac_new_detailed}
\end{table}

\begin{table}[!htbp]
	\centering
	\caption{
		Details of the GPU servers used for model training and inference.
		Best viewed under high magnification.
	}
	\begin{tabular}{c}
		\includegraphics[width=0.48\textwidth]{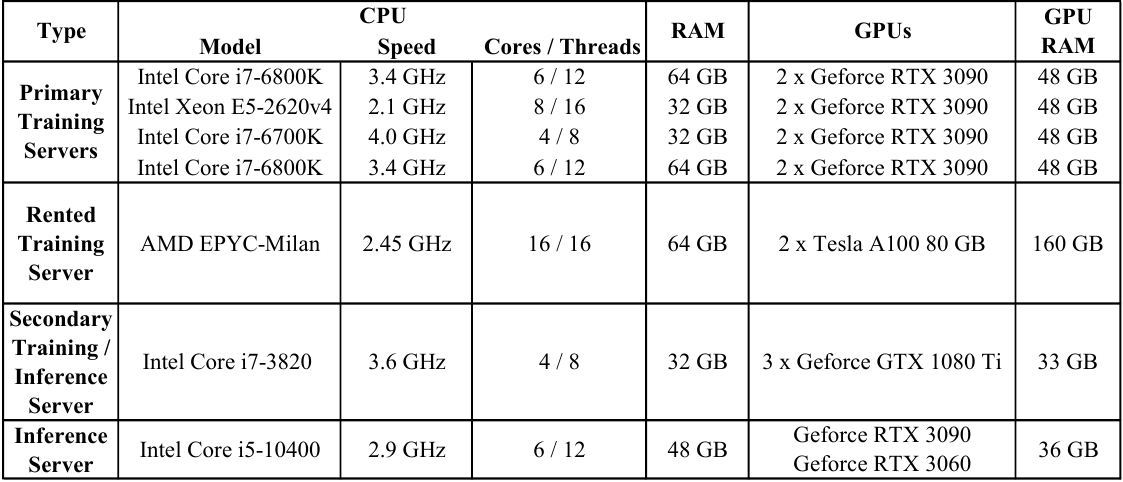}
	\end{tabular}	
	\label{tab:servers}
\end{table}

\begin{table*}[!htbp]
	\centering
	\caption{
		Details of the models whose results are reported in Sec. \ref{res_overview} except a few models in Sec. \ref{vid_det_detrac} whose details are in Table \ref{tab:detrac_new_detailed}.
	}
	\begin{tabular}{c}
		\includegraphics[width=\textwidth]{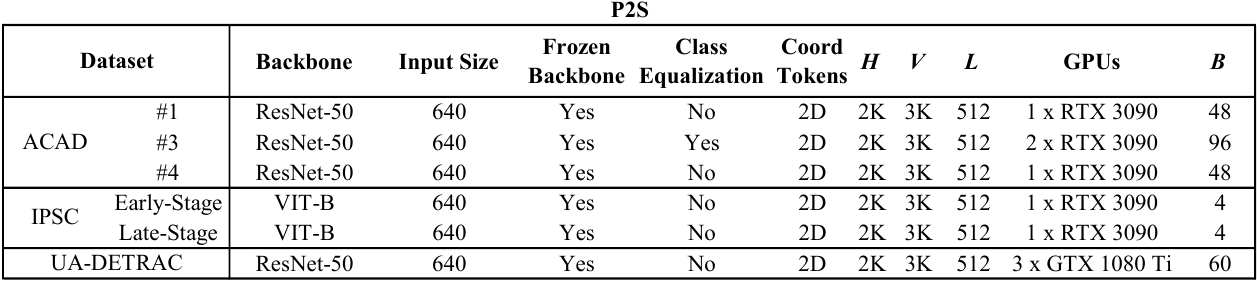}\\
		\includegraphics[width=\textwidth]{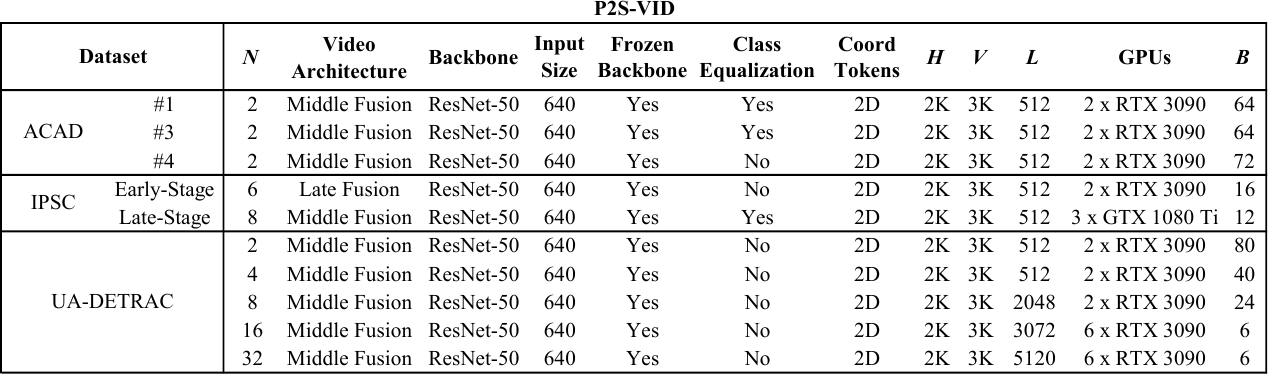}
	\end{tabular}	
	\label{tab:model_configs}
\end{table*}

\bibliographystyle{IEEEtran}
\bibliography{all_references.bib}

\begin{IEEEbiography}[{\includegraphics[width=1in,height=1.25in,clip,keepaspectratio]{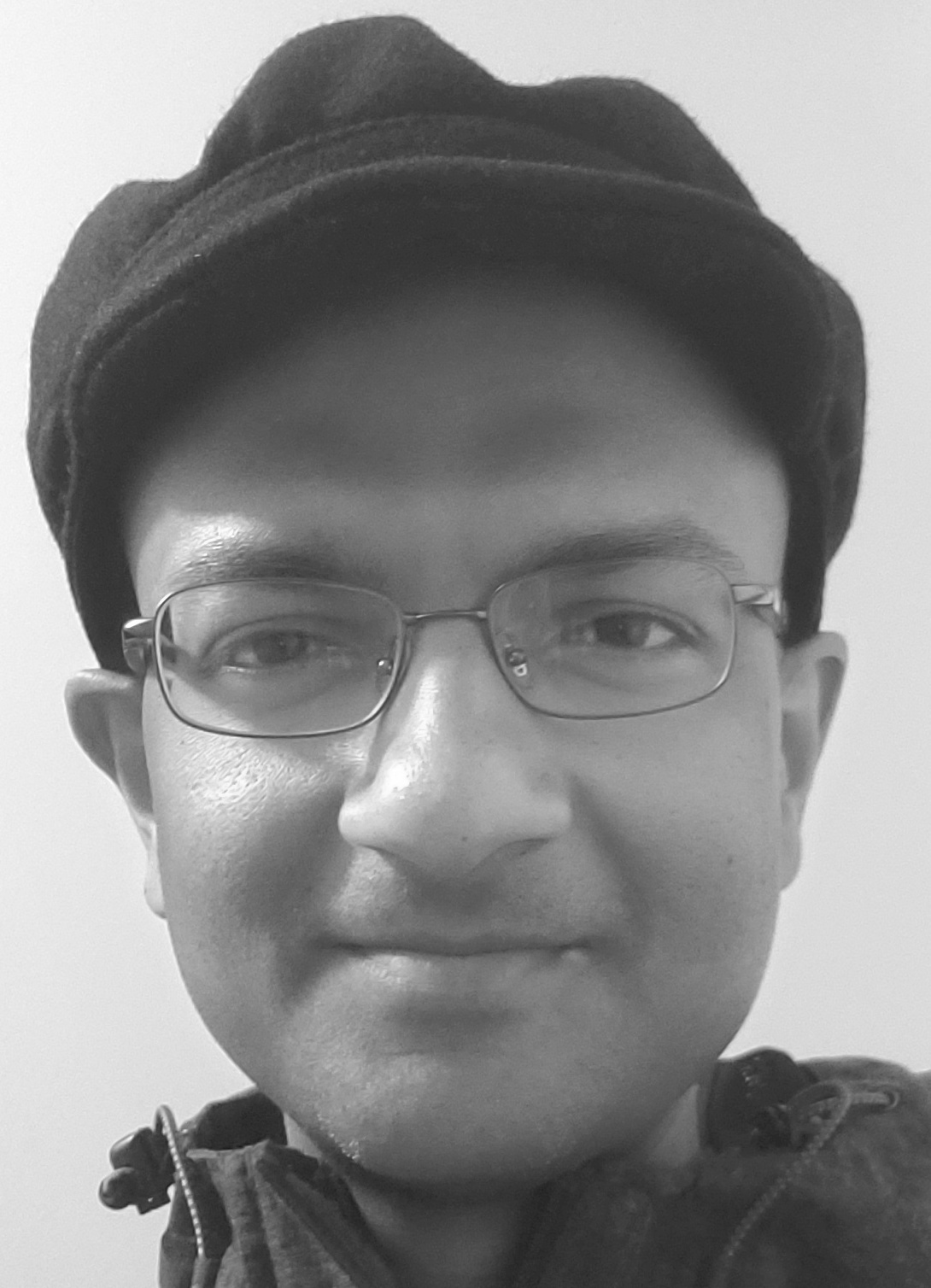}}]{Abhineet Singh}  received the B.Tech. degree in information technology from IIIT Allahabad, Prayagraj, India, in 2013, the M.Sc. degree in computing science from the University of Alberta,	Edmonton, AB, Canada, in 2017 and the Ph.D. degree from the same department in 2025. 
	
He is currently working as a machine learning developer for an Edmonton-based agricultural automation company named Mojow Autonomous Solutions.
His research interests include computer vision and machine learning in general and application of deep learning for object detection, tracking, and segmentation in particular.
\end{IEEEbiography}

\begin{IEEEbiography}[{\includegraphics[width=1in,height=1.25in,clip,keepaspectratio]{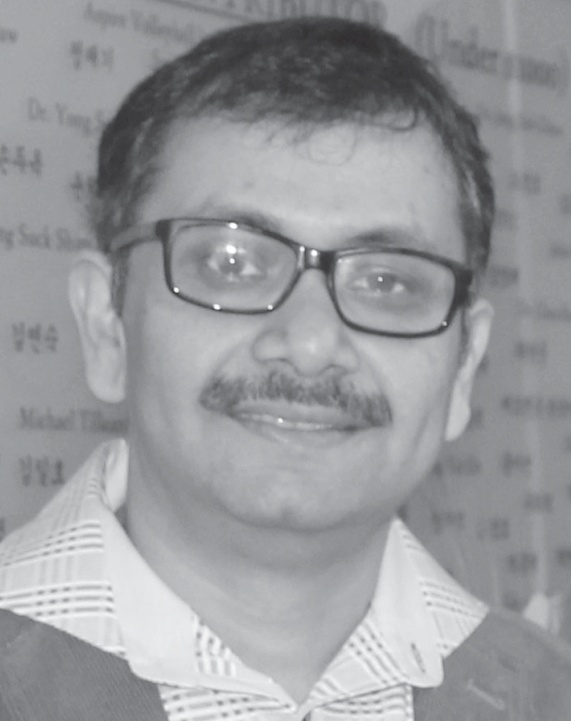}}]{Nilanjan Ray} received Bachelor of Mechanical Engineering from Jadavpur University, Calcutta, India, in 1995, M.Tech. in Computer Science from Indian Statistical Institute, Calcutta, in 1997, and Ph.D. in Electrical Engineering from the University of Virginia, USA, in 2003. After having two years of postdoctoral research and a year of industrial work experience he joined the Department of Computing Science, University of Alberta in 2006, where, currently, he is a full professor.

Nilanjan's research is in computer vision, image analysis and visual recognition with deep learning. He is interested in medical imaging and general computer vision applications including classification, recognition, semantic segmentation, object tracking, image registration and motion detection. He has published over 150 articles in these areas. 

Nilanjan's professional activities include serving as a General Co-chair for AI/GI/CRV conference in 2017, Associate Editor for IEEE Transactions on Image Processing (2013-2017), IET Image Processing (2016-2021), BMVC 2022 PC member, Editorial board member of Discover Imaging, 2024-.
\end{IEEEbiography}

\EOD

\end{document}